\documentclass[journal]{IEEEtran}

\usepackage{times}
\usepackage{epsfig}
\usepackage{amsmath}
\usepackage{amssymb}
\usepackage{multirow}
\usepackage[]{booktabs}
\usepackage{graphicx}
\usepackage{subcaption}
\usepackage{float}
\usepackage[colorlinks]{hyperref}
\usepackage{cite}

\begin{document}

\title{Exploiting Temporal Contexts with Strided Transformer for 3D Human Pose Estimation}

\author{
Wenhao Li, Hong Liu$^{\dagger}$, Runwei Ding, Mengyuan Liu, Pichao Wang, and Wenming Yang
\thanks{$^{\dagger}$ Corresponding author.}
\thanks{W. Li, H. Liu, and R. Ding are with Key Laboratory of Machine Perception, Shenzhen Graduate School, Peking University, Beijing 100871, China. 
E-mail: \{wenhaoli, hongliu, dingrunwei\}@pku.edu.cn. 
M. Liu is with School of Intelligent Systems Engineering, Sun Yat-sen University, China. E-mail: nkliuyifang@gmail.com. 
P. Wang is with Alibaba Group, Bellevue, WA, 98004, USA. 
E-mail: pichao.wang@alibaba-inc.com. 
W. Yang is with the Shenzhen Key Lab of Information Science and Technology, Shenzhen Engineering Lab of
IS\&DRM, Department of Electronic Engineering, Graduate School at Shenzhen, Tsinghua University, Shenzhen 518055, China. 
E-mail: yang.wenming@sz.tsinghua.edu.cn

This work is supported by National Key R\&D Program of China (No. 2020AAA0108904), Basic and Applied Basic Research Foundation of Guangdong (No. 2020A1515110370), Science and Technology Plan of Shenzhen (Nos. JCYJ20190808182209321, JCYJ20200109140410340).}
}

% \markboth{IEEE Transactions on Multimedia}
\markboth{}
{Li \MakeLowercase{\textit{et al.}}: 
Exploiting Temporal Contexts with Strided Transformer for 3D Human Pose Estimation}

\maketitle

\begin{abstract}
Despite the great progress in 3D human pose estimation from videos, it is still an open problem to take full advantage of a redundant 2D pose sequence to learn representative representations for generating one 3D pose. To this end, we propose an improved Transformer-based architecture, called Strided Transformer, which simply and effectively lifts a long sequence of 2D joint locations to a single 3D pose. Specifically, a Vanilla Transformer Encoder (VTE) is adopted to model long-range dependencies of 2D pose sequences. To reduce the redundancy of the sequence, fully-connected layers in the feed-forward network of VTE are replaced with strided convolutions to progressively shrink the sequence length and aggregate information from local contexts. The modified VTE is termed as Strided Transformer Encoder (STE), which is built upon the outputs of VTE. STE not only effectively aggregates long-range information to a single-vector representation in a hierarchical global and local fashion, but also significantly reduces the computation cost. Furthermore, a full-to-single supervision scheme is designed at both full sequence and single target frame scales applied to the outputs of VTE and STE, respectively. This scheme imposes extra temporal smoothness constraints in conjunction with the single target frame supervision and hence helps produce smoother and more accurate 3D poses. The proposed Strided Transformer is evaluated on two challenging benchmark datasets, Human3.6M and HumanEva-I, and achieves state-of-the-art results with fewer parameters. Code and models are available at \url{https://github.com/Vegetebird/StridedTransformer-Pose3D}. 
\end{abstract}


\IEEEpeerreviewmaketitle

\section{Introduction}
\IEEEPARstart{3}{D} human pose estimation is a classic computer vision task that aims to estimate 3D joint locations of a human body from images or videos. 
This task has drawn tremendous attention in the past decades~\cite{radwan2013monocular,li20143d,zhao20183,hu20213dbodynet} since it plays a significant role in wide applications, such as clinic~\cite{kadkhodamohammadi2021generalizable}, computer animation~\cite{pullen2002motion}, action recognition~\cite{wang2018depth,liu2017robust,liu2018recognizing,wei2019learning,song2021constructing,chen2021learning,li2021memory,yang2021unik,zhang2017action,chen2017multi}, and human-robot interaction~\cite{garcia2019human,gui2018teaching}. 
Many state-of-the-art approaches adopt a two-stage pipeline (\emph{i.e.}, 2D-to-3D lifting method)~\cite{martinez2017simple,pavllo20193d,hua2021weakly}, which first estimates 2D keypoints and then lifts them to 3D space. 
Although the 2D-to-3D lifting method benefits from the reliable performance of 2D pose detectors, it is still a highly ill-posed problem due to the inherent ambiguity in depth, since multiple 3D interpretations can be projected to the same 2D pose in the image space. 

To alleviate this problem, temporal context information has been investigated by many researchers. 
Some methods~\cite{lee2018propagating,rayat2018exploiting,cai2019exploiting} leverage past and future data in the sequence to predict the 3D pose of the target frame. 
For instance, Cai \emph{et al.}~\cite{cai2019exploiting} presented a local-to-global graph convolutional network to exploit spatio-temporal relations to estimate 3D keypoints from a 2D pose sequence.
However, these approaches have small temporal receptive fields and limited temporal correlation windows, thus suffering from modeling long-range dependencies. 

\begin{figure}[t]
   \centering
   \includegraphics[width=1.00 \linewidth]{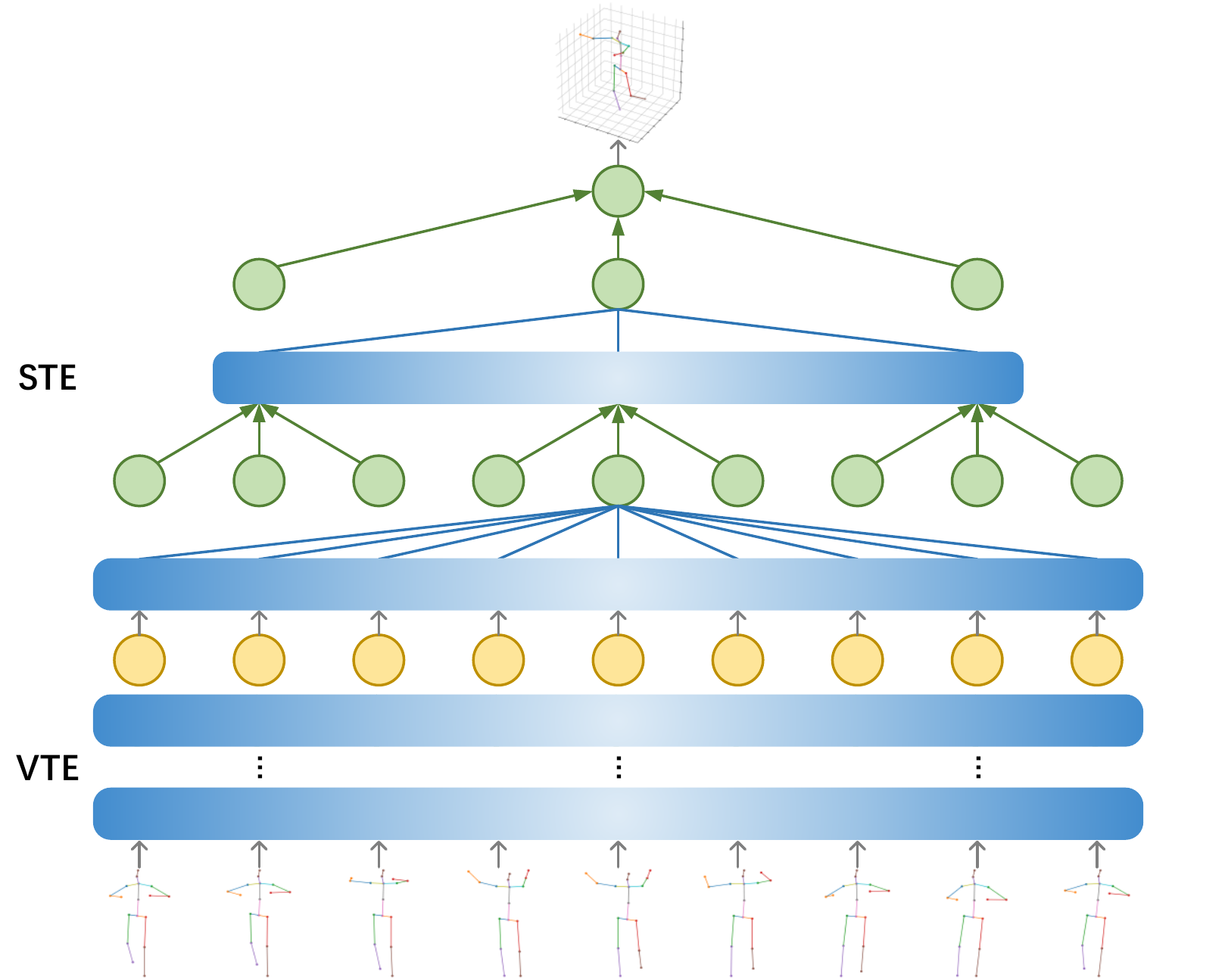}
   \caption
   {
      Our Strided Transformer Encoder (STE) takes the outputs of Vanilla Transformer Encoder (VTE) as input (yellow) and generates a 3D pose for the target frame as output (top). 
      The self-attention mechanism (blue) concentrates on global context and the strided convolution (green) aggregates information from local contexts. 
   }
   \label{fig:moti}
\end{figure}

\begin{figure*}[t]
   \centering
   \includegraphics[width=1.00 \linewidth]{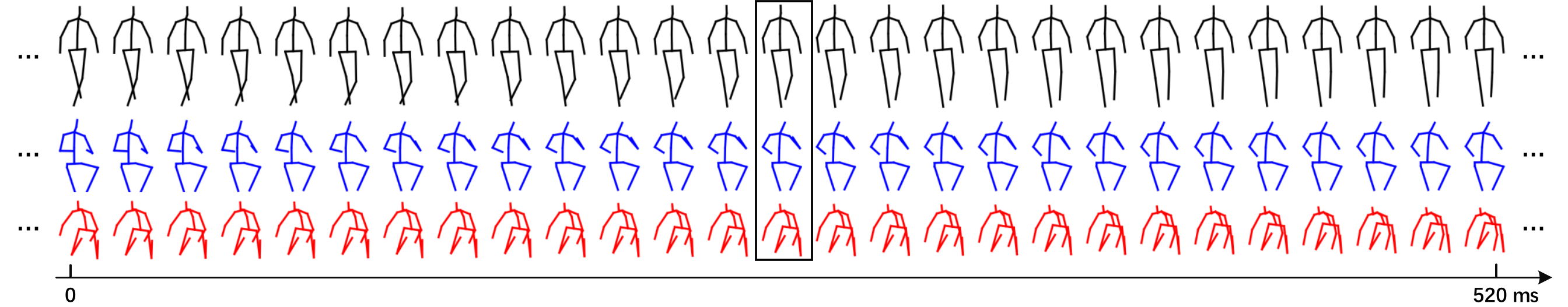}
   \caption
   {
      Example of 2D pose sequences of 27 consecutive frames (520 ms) on Human3.6M dataset (captured from 50 Hz cameras). 
      It contains huge redundant information as nearby poses are same. 
      The rectangle denotes the center frame. 
   }
   \label{fig:2d_pose}
\end{figure*}

Vanilla Transformer~\cite{Attention} is developed for exploiting long-range dependencies and achieves tremendous success in natural language processing~\cite{tay2020efficient,zihang2020funnel-transformer} and computer vision~\cite{han2020survey,he2021transreid,li2021trear,han2020exploiting,li2021transformer}. 
It consists of a self-attention module and a position-wise feed-forward network (FFN). 
The self-attention module computes pairwise dot-product among all input elements to capture global-context information, and the FFN acts as pattern detectors over the input across all layers~\cite{geva2020transformer}. 
Such a design looks like a good choice for the 2D-to-3D pose lifting method to capture long-range dependencies. 
However, there are several shortcomings in the Vanilla Transformer Encoder (VTE)~\cite{Attention}:
(i) The full-length sequence in the forward pass across all layers contains significant redundancy for video-based pose estimation as nearby poses are quite similar, as illustrated in Fig.~\ref{fig:2d_pose}. 
(ii) The time and memory complexity of the attention operation grows quadratically with the input length, making it very expensive to process long sequences. 
Thus, the receptive field may be forced to decrease in real-time applications, whereas a large receptive field is important to enhance the estimation consistency~\cite{liu2020attention}. 
(iii) The VTE architecture is less capable to extract fine-grained local feature patterns, which is well-known to be crucial for computer vision tasks. 
To mitigate these issues, we propose to gradually merge nearby poses to shrink the sequence length until one representation of the target pose is acquired. 
An alternative is to perform pooling operation after the FFN~\cite{zihang2020funnel-transformer}. 
However, lots of valuable information will be lost using pooling operation, and the local information can not be well exploited. 
Motivated by the previous methods~\cite{pavllo20193d,liu2020attention} that are able to elegantly handle variable-length sequences via temporal convolutions, we propose to replace fully-connected layers in FFN with strided convolutions to progressively reduce the sequence length. 
The modified Transformer is dubbed Strided Transformer Encoder (STE), as shown in Fig.~\ref{fig:moti}. 
With the proposed STE, we can model both global and local information in a hierarchical architecture, and the computation in FFN can be traded off for constructing a deeper model to boost the model capacity. 

Although the STE can aggregate long-range information to a single-pose representation, it remains a question whether this single representation is enough to represent a long sequence and how to make this representation work in improving the performance. 
We observe that directly supervising the model at a single target frame scale always breaks temporal smoothness among video frames, while only supervising at a full sequence scale cannot explicitly learn a specific representation for the target frame. 
These observations encourage us to develop a method that can effectively embed both scales into a learnable framework. 
Therefore, based on the outputs of VTE and STE, a full-to-single supervision scheme is designed at both full and single scales, which can impose extra temporal smoothness constraints at the full sequence scale and refine the estimation at the single target frame scale. 
This scheme brings great benefits in producing smoother and more accurate 3D poses. 

\begin{figure*}[t]
   \centering
   \includegraphics[width=1.00 \linewidth]{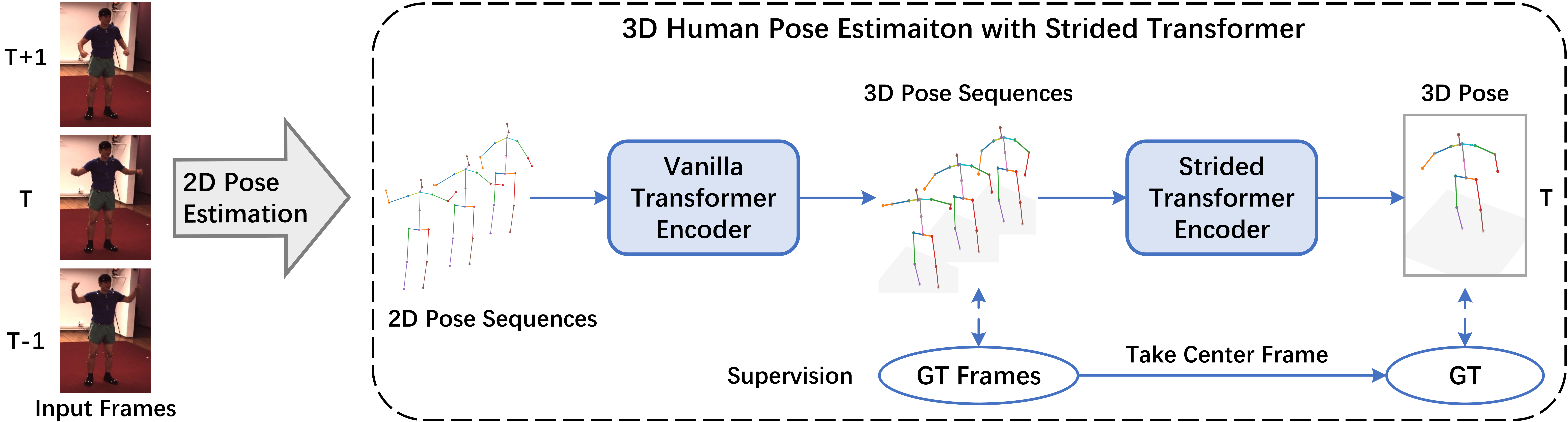}
   \caption
   {
      Overview of our proposed Strided Transformer for predicting the 3D joint locations of the target frame (center frame) from the estimated 2D pose sequences. 
      It mainly consists of a Vanilla Transformer Encoder (VTE) and a Strided Transformer Encoder (STE). 
      The network first models long-range information via VTE and then aggregates the information into one target pose representation from the proposed STE. 
      The model is trained end-to-end at both full sequence and single target frame scales. 
   }
   \label{fig:pipline}
\end{figure*}

The proposed architecture is called Strided Transformer, as shown in Fig.~\ref{fig:pipline}. 
Extensive experiments are conducted on two standard 3D human pose estimation datasets, \emph{i.e.}, Human3.6M~\cite{ionescu2013human3} and HumanEva-I~\cite{sigal2010humaneva}. 
Experimental results show that the proposed method achieves state-of-the-art performance. 

Our contributions are summarized as follows:
\begin{itemize}
   \item We propose a new Transformer-based architecture for 3D human pose estimation called Strided Transformer, which can simply and effectively lift a long 2D pose sequence to a single 3D pose. 

   \item To reduce the sequence redundancy and computation cost, Strided Transformer Encoder (STE) is introduced to gradually reduce the temporal dimensionality and aggregate long-range information into a single-vector representation of pose sequences in a hierarchical global and local fashion. 
   \item A full-to-single supervision scheme is designed to impose extra temporal smoothness constraints during training at the full sequence scale and further refine the estimation at the single target frame scale. 

   \item State-of-the-art results are achieved with fewer parameters on two commonly used benchmark datasets, making our method a strong baseline for Transformer-based 3D pose estimation. 
\end{itemize}

\section{Related Work}
At the early stage of applying deep neural networks on 3D pose estimation task, many methods~\cite{pavlakos2017coarse,sun2018integral,zhao2019semantic,liu2019feature} learned the direct mapping from RGB images to 3D poses (\emph{i.e.}, one-stage pose estimation).
However, these methods require sophisticated architectures with high computation costs, which are impractical in realistic applications.  

\textbf{Two-stage pose estimation.}
Two-stage methods formulate the problem of 3D human pose estimation as 2D keypoint detection followed by 2D-to-3D lifting estimation~\cite{martinez2017simple,fang2018learning,xu2021graph}. 
Recent works show that 3D locations of body joints can be efficiently and effectively recovered using detected 2D poses from state-of-the-art 2D pose detectors, and this 2D-to-3D pose lifting method outperforms one-stage approaches.  
For example, 
%A simple yet effective baseline for 3d human pose estimation
Martinez \emph{et al.}~\cite{martinez2017simple} lifted 2D joint locations to 3D space via a fully-connected residual network. 
% Learning Pose Grammar to Encode Human Body Configuration for 3D Pose Estimation
Fang \emph{et al.}~\cite{fang2018learning} proposed a pose grammar model to encode the human body configuration of human poses from 2D space to 3D space. 
% PoseAug: A Differentiable Pose Augmentation Framework for 3D Human Pose Estimation
To improve the generalization of the trained 2D-to-3D pose estimator, Gong~\cite{gong2021poseaug}
introduced a pose augmentation framework that is differentiable. 
We also follow this two-stage pipeline because it is widely adopted among the state-of-the-art methods in this domain. 

\textbf{Video pose estimation.}
Recently, many approaches tried to exploit temporal information~\cite{rayat2018exploiting,pavllo20193d,cai2019exploiting,wang2020motion} to improve the accuracy and the smoothness of the estimated 3D pose sequence. 
% Exploiting Temporal Information for 3D Human Pose Estimation
To predict temporally consistent 3D poses, Hossain \emph{et al.}~\cite{rayat2018exploiting} designed a sequence-to-sequence network with LSTM. 
%3D human pose estimation in video with temporal convolutions and semi-supervised training
Pavllo \emph{et al.}~\cite{pavllo20193d} introduced a fully convolutional model based on dilated temporal convolutions. 
Cai \emph{et al.}~\cite{cai2019exploiting} directly chose the 3D pose of the target frame from the outputs of the proposed graph-based method and then fed it to a refinement model. 
% Motion Guided 3D Pose Estimation from Videos
To produce smoother 3D sequences, Wang \emph{et al.}~\cite{wang2020motion} designed an U-shaped graph convolutional network and involved motion modeling into learning. 
However, the temporal connectivity of these architectures is inherently limited and is mainly constrained to simple sequential correlations. 
Different from most existing works that employed LSTM-based~\cite{rayat2018exploiting}, graph-based~\cite{cai2019exploiting,wang2020motion}, or temporal convolutional networks~\cite{pavllo20193d,liu2020attention,chen2021anatomy} to exploit temporal information, we propose a Transformer-based architecture to capture long-range dependencies from input 2D pose sequences. 
Furthermore, compared with previous methods~\cite{cai2019exploiting,wang2020motion} that either utilize a refinement model or use a motion loss to improve estimations, we design a full-to-single supervision scheme that refines the intermediate predictions to produce smoother and more accurate estimations. 

\textbf{Visual Transformers.}
Transformer models first proposed in~\cite{Attention} are commonly used in various language tasks. 
Recently, Transformers have shown promising performance in many computer vision tasks, such as object detection~\cite{carion2020end,zhu2020deformable} and image classification~\cite{dosovitskiy2020image,yuan2021tokens}. 
DETR~\cite{carion2020end} presented a new Transformer-based design for object detection systems. 
ViT~\cite{dosovitskiy2020image} proposed to apply a standard Transformer architecture directly to sequential image patches for image classification. 
METRO~\cite{lin2020end} introduced a Transformer framework to reconstruct 3D human pose and mesh from a single image. 
However, METRO focused on the one-stage pose estimation and ignores the temporal information across frames. 
Unlike DETR~\cite{carion2020end}, ViT~\cite{dosovitskiy2020image}, or METRO~\cite{lin2020end} that directly apply Transformer to images, we utilize a Transformer-based architecture to effectively map 2D keypoints to 3D poses. 
Additionally, efficient strided convolutions are incorporated into Transformer models to address the redundancy problem for the video-based 3D pose estimation task. 

\section{Method}
In this section, we first present an overview of the proposed Strided Transformer for 3D human pose estimation from a 2D video stream, and then show how our Transformer-based architecture learns a representative single-pose representation from redundant sequences resulting in an enhanced estimation. 
Finally, the complexity analysis of our network is presented. 

\subsection{Overview}
The overall framework of our proposed method is illustrated in Fig.~\ref{fig:pipline}. 
Given a sequence of the estimated 2D poses
$P=\left\{p_{1}, \ldots, p_{T}\right\}$ from videos, we aim at reconstructing 3D joint locations $X \in \mathbb{R}^{J \times 3}$ for a target frame (center frame), where $p_{t} \in \mathbb{R}^{J \times 2}$ denotes the 2D joint locations at frame $t$, $T$ is the number of video frames, and $J$ is the number of joints. 
The network contains a Vanilla Transformer Encoder (VTE) followed by a Strided Transformer Encoder (STE), which is trained in a full-to-single prediction scheme at both full sequence and single target frame scales. 
Specifically, VTE is first used to model long-range information and is supervised by the full sequence scale to enforce temporal smoothness. 
Then, the proposed STE aggregates the information to generate one target pose representation and is supervised by the single target frame scale to produce more accurate estimations.  

\subsection{Strided Transformer Encoder}
\label{sec:STE}
Despite the substantial performance gains achieved by Transformers~\cite{Attention} in many computer vision tasks, the full-length token representation makes it unsuitable for many video-based vision tasks that only require a single-vector representation of a sequence. 
To this end, STE is proposed to gradually compress the sequence of hidden states and model both global and local information in a hierarchical architecture. 
Each layer of the proposed STE consists of a multi-head self-attention (MSA) and a convolutional feed-forward network (CFFN). 

\subsubsection{Multi-head self-attention}
The core mechanism of the Transformer model is MSA~\cite{Attention}. 
Suppose there are a set of queries ($Q$), keys ($K$), and values ($V$) of dimension $d_{m}$. 
Then the MSA can be computed as:
\begin{align}
   head_{i}=\operatorname{Self-Attn}\left(Q W_{i}^{Q}, K W_{i}^{K}, V W_{i}^{V}\right), \\
   \operatorname{MSA}(Q, K, V)=\operatorname{Concat}\left(head_{1}, \ldots, head_{h}\right) W^{O},
\end{align}
where $\operatorname{Self-Attn}(Q, K, V)=\operatorname{softmax}\left(Q K^{T} / \sqrt{d_{k}}\right) V$ and $W_{i}^{Q} \in \mathbb{R}^{d_{m} \times d_{k}}, W_{i}^{K} \in \mathbb{R}^{d_{m} \times d_{k}}, W_{i}^{V} \in \mathbb{R}^{d_{m} \times d_{v}}$, and $W^{O} \in \mathbb{R}^{h d_{v} \times d_{m}}$ are parameter matrices. 
The hyperparameter $h$ is the number of multi-attention heads, $d_{m}$ is the dimension of the model, and $d_{k} = d_{v} = d_{m}/h$ in our implementation. 

\subsubsection{Convolutional feed-forward network}
In the existing fully-connected (FC) layers in the FFN of VTE (Eq. (\ref{equ:ffn})), it always maintains a full-length sequence of hidden representations across all layers with a high computation cost. 
It contains significant redundancy for video-based pose estimation, as nearby poses are quite similar. 
However, to reconstruct more accurate 3D body joints of the target frame, crucial information should be extracted from the entire pose sequences. 
Therefore, it requires selectively aggregating useful information. 

To tackle this issue, inspired by the previous works~\cite{pavllo20193d,liu2020attention} that employ temporal convolutions to effectively shrink the sequence length, we make modifications to the generic FFN. 
Given the input feature vector $Z \in \mathbb{R}^{T \times D_{in}}$ with $T$ sequences and $D_{in}$ channels to generate an output of $(\tilde{T}, D_{out})$ features, the operation performed by FC in FFN can be formulated as: 
\begin{equation}
   \operatorname {FC}_{t, d_{out}}(z)= \sum_{i}^{D_{i n}} w_{d_{o u t}, i} * z_{t, i} \; .
   \label{equ:ffn}
\end{equation}

If 1D convolution is considered with kernel size $K$ and strided factor $S$, a strided convolution in CFFN can be computed as:
\begin{equation}
   \operatorname {Conv}_{S(t), c_{out}}(z)=\sum_{i}^{D_{i n}} \sum_{k}^{K} w_{d_{out}, i, k} * z_{S(t-\frac{K-1}{2}+k), i} \; . 
\end{equation}

In this way, fully-connected layers in FFN of VTE are replaced with strided convolutions. 
The modified VTE is termed as Strided Transformer Encoder (STE), which can be represented as:
\begin{align}
   \label{equ:ste_1}
   & \hat{Z}^{n-1}=Z^{n-1}+\operatorname{MSA}(\operatorname{LN}(Z^{n-1})), \\
   & Z^{n}=\operatorname{MaxPool}(\hat{Z}^{n-1})+\operatorname{CFFN}(\operatorname{LN}(\hat{Z}^{n-1})),
   \label{equ:ste_2}
\end{align}
where $\operatorname{LN}(\cdot)$ denotes the layer normalization, $\operatorname{MaxPool}(\cdot)$ denotes the max pooling operation, and $n \in[1, \ldots, N]$ is the index of STE layers. 

\begin{figure}[t]
   \centering
   \includegraphics[width=1.00 \linewidth]{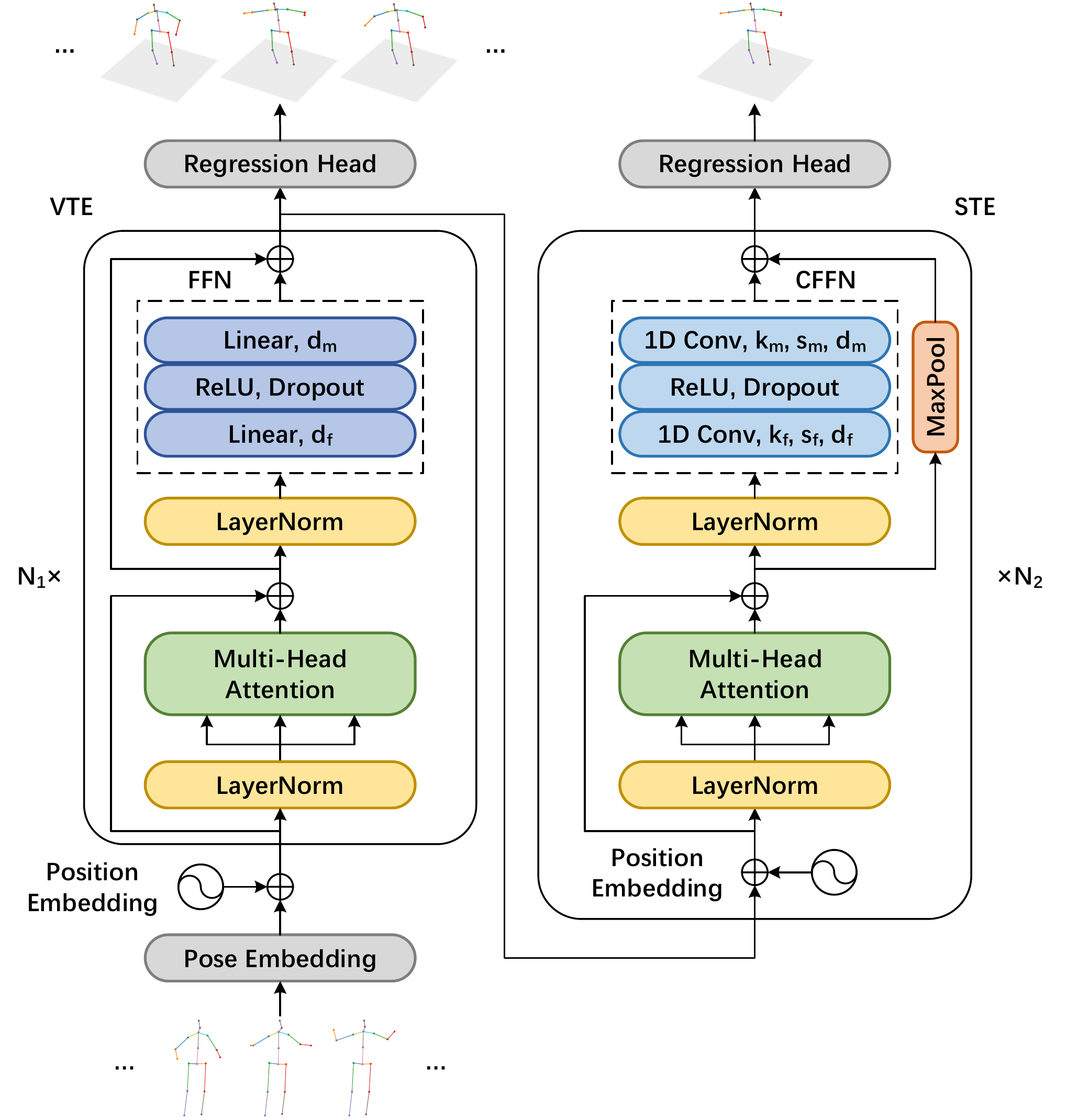}
   \caption
   {
      The network architecture of our proposed Strided Transformer. 
      The left is the VTE and the right is the STE. 
      Here, $N_{1}$ and $N_{2}$ denote the number of layers of the two modules, respectively. 
      The hyperparameters $k$, $s$, $d_{m}$ and $d_{f}$ are the kernel size, the strided factor, the dimension, and the number of hidden units. 
      The max pooling operation is applied to the residuals to match the temporal dimensions. 
   }
   \label{fig:network}
\end{figure}

The STE is a hierarchical global and local architecture, where the self-attention mechanism models global context and the strided convolution helps capture local contexts, as presented in Fig.~\ref{fig:network} (right). 
It gradually merges the nearby poses to a short sequence length representation, as illustrated in Fig.~\ref{fig:architecture}. 
Importantly, through such a hierarchical design, the redundancy of the sequence and the computation cost can be reduced. 

\subsection{Network Architecture}
\label{sec:transpose}
\begin{figure*}[t]
   \centering
   \includegraphics[width=1.00 \linewidth]{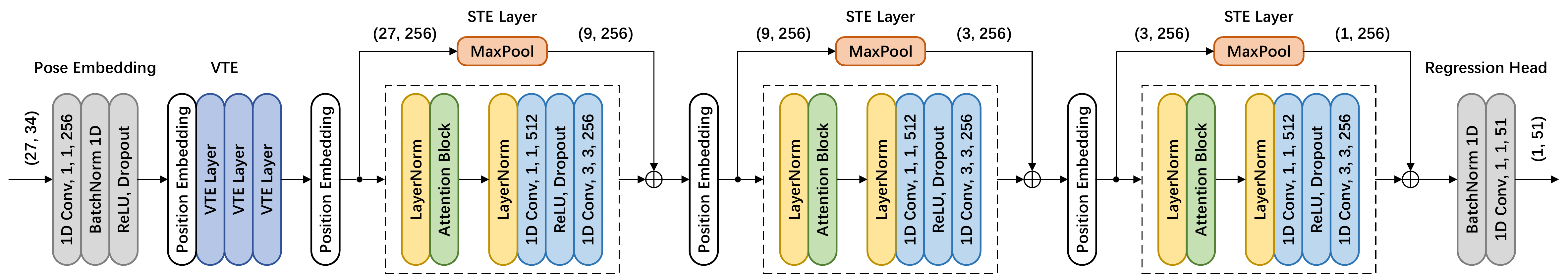}
   \caption
   {
      An instantiation of the proposed Strided Transformer network.  
      It reconstructs the target 3D body joints by progressively reducing the sequence length. 
      The input consists of 2D keypoints for a receptive field of 27 frames with $J = 17$ joints. 
      Convolutional feed-forward networks are in blue where $(3, 3, 256)$ denotes kernels of size 3 with strided factor 3 and 256 output channels. 
      The tensor sizes are shown in parentheses, \emph{e.g.}, $(27, 34)$ denotes 27 frames and 34 channels. 
      Due to strided convolutions, the max pooling operation is applied to the residuals to match the shape of subsequent tensors.
   }
   \label{fig:architecture}
\end{figure*}

In this section, we describe how to use the proposed Transformer-based network architecture to estimate 3D human poses from a sequence of 2D poses. 
As shown in Fig.~\ref{fig:architecture}, the proposed network is composed of four components: a pose embedding, a Vanilla Transformer Encoder (VTE), a Strided Transformer Encoder (STE), and a regression head. 

\subsubsection{Pose embedding}
Given a sequence of the estimated 2D poses $P \in \mathbb{R}^{T \times J \times 2}$, the pose embedding first concatenates $(x, y)$ coordinates of the $J$ joints for each frame to tokens $P^{\prime} \in \mathbb{R}^{T \times (J \cdot 2)}$, and then embeds each token to a high dimensional feature $Z_{0} \in \mathbb{R}^{T \times d_{m}}$ using a 1D convolutional layer with $d_{m}$ channels, followed by batch normalization, dropout, and a ReLU activation. 

\subsubsection{Vanilla Transformer Encoder}
Suppose that the VTE consists of $N_{1}$ layers, the learnable position embedding $E_{1} \in \mathbb{R}^{T \times d_{m}}$ is used before the first layer of VTE, which can be formulated as follows: 
\begin{equation}
   Z_{1}^{0} = Z_{0} + E_{1}. 
\end{equation}

Then, given the embedded feature $Z_{1}^{0}$, the VTE layers can be represented as: 
\begin{align}
   \hat{Z}_{1}^{n-1}&=Z_{1}^{n-1}+\operatorname{MSA}(\operatorname{LN}(Z_{1}^{n-1})), \\
   Z_{1}^{n}&=\hat{Z}_{1}^{n-1}+\operatorname{FFN}(\operatorname{LN}(\hat{Z}_{1}^{n-1})),
\end{align}
where $n \in[1, \ldots, N_{1}]$ is the index of VTE layers. 
It can be expressed by using a function of a VTE layer $\operatorname{VTE}(\cdot)$: 
\begin{equation}
   Z_{1}^{n} = \operatorname{VTE}(Z_{1}^{n-1}).  
\end{equation}

\subsubsection{Strided Transformer Encoder}
For the STE, it is built upon the outputs of VTE and takes the $Z_{1}^{N_{1}} \in \mathbb{R}^{T \times d_{m}}$ as input. 
The learnable position embeddings $E_{2} \in \mathbb{R}^{S(t) \times d_{m}}$ with strided factor $S$ are used for every layer of STE due to the different sequence lengths. 
Then, the STE layers can be represented as follows:
\begin{equation}
   Z_{2}^{n} = \operatorname{STE}(Z_{2}^{n-1} + E_{2}^{n}), 
\end{equation}
where $n \in[1, \ldots, N_{2}]$ is the index of STE layers, $Z_{2}^{0} = Z_{1}^{N_{1}}$, and $\operatorname{STE}(\cdot)$ denotes the function of an STE layer whose details can be found in Eq. (\ref{equ:ste_1}) and Eq. (\ref{equ:ste_2}). 

\subsubsection{Regression head}
In order to perform the regression, a batch normalization and a 1D convolutional layer are applied to the outputs of VTE and STE, $Z_{1}^{N_{1}} \in \mathbb{R}^{T \times d_{m}}$ and $Z_{2}^{N_{2}} \in \mathbb{R}^{1 \times d_{m}}$, respectively. 
Finally, the outputs of 3D pose prediction are $\tilde{X}$ and $X$, where $\tilde{X} \in \mathbb{R}^{T \times J \times 3}$ and $X \in \mathbb{R}^{J \times 3}$ are predictions of the 3D pose sequence and the 3D joint locations of the target frame, respectively. 

\subsection{Full-to-Single Prediction}
The iterative refinement scheme, aimed at producing predictions in multiple processing stages, is effective for 3D pose estimation~\cite{pavlakos2017coarse,cai2019exploiting}. 
Motivated by the success of such iterative processing, we also consider a refinement scheme. 
A full-to-single scheme is proposed to incorporate both full sequence and single target frame scales constraints into the framework. 
This scheme further refines the intermediate predictions to produce more accurate estimations rather than using a single component with a single output. 
More precisely, the full sequence scale can enforce temporal smoothness and the single target frame scale helps learn a specific representation for the target frame. 

\subsubsection{Full sequence scale}
The first step is to supervise at full sequence scale by imposing extra temporal smoothness constraints during training from the output of VTE followed by a regression head. 
A sequence loss $\mathcal{L}_{f}$ is used to improve upon single frame predictions for temporal consistency over a sequence. 
This loss ensures that the estimated 3D pose sequences $\tilde{X} \in \mathbb{R}^{T \times J \times 3}$ coincide with the ground truth 3D joint sequences $Y \in \mathbb{R}^{T \times J \times 3}$: 
\begin{equation}
   \mathcal{L}_{f}=\sum_{t=1}^{T} \sum_{i=1}^{J}\left\|Y_{i}^{t}-\tilde{X}_{i}^{t}\right\|_{2},
\end{equation}
where $\tilde{X}_{i}^{t}$ and $Y_{i}^{t}$ represent the sequence of estimated 3D poses and ground truth 3D joint locations of joint $i$ at frame $t$, respectively. 

\subsubsection{Single target frame scale}
In the second step, the supervision is adopted on the output of STE followed by a regression head. 
A single-frame loss $\mathcal{L}_{s}$ is used to refine the estimation at the single target frame scale. 
It minimizes the distance between the estimated 3D pose $X \in \mathbb{R}^{J \times 3}$ and the target ground truth 3D joint annotation $Y \in \mathbb{R}^{J \times 3}$: 
\begin{equation}
   \mathcal{L}_{s}=\sum_{i=1}^{J}\left\|Y_{i}-X_{i}\right\|_{2},
\end{equation}
where ${X}_{i}$ and $Y_{i}$ represent the target frame's estimated 3D pose and ground truth 3D joint locations of joint $i$, respectively. 

\subsubsection{Loss function}
In our implementation, the model is supervised at both full sequence scale and single target frame scale. 
We train the entire network in an end-to-end manner with the total loss:
\begin{equation}
   \mathcal{L}=\lambda_{f}\mathcal{L}_{f}+\lambda_{s}\mathcal{L}_{s},
\end{equation}
where $\lambda_{f}$ and $\lambda_{s}$ are weighting factors. 

\subsection{Complexity Analysis}
In this section, we use floating-point operations (FLOPs) to measure the computational cost and analyze the compression ratio of our proposed Strided Transformer network. 
Given the sequence length $t$, dimension $d_{m}=d_{f}/2=d$, strided factor $s$, and kernel size $k$, the FLOPs of a VTE layer $\mathcal{F}_{VTE}^{n}$ and an STE layer $\mathcal{F}_{STE}^{n}$ can be computed by: 
\begin{equation}
\begin{aligned}
   \mathcal{F}_{VTE}^{n}(t, d) &=\mathcal{F}_{MSA}^{n}(t, d)+\mathcal{F}_{FFN}^{n}(t, d) \\
   &=8 t d^{2} + 2 t^{2} d,
\end{aligned}
\end{equation}
\begin{equation}
\begin{aligned}
   \mathcal{F}_{STE}^{n}(t, d, s) &=\mathcal{F}_{MSA}^{n}(t, d, s)+\mathcal{F}_{CFFN}^{n}(t, d, s) \\
   &=(6+2s^{-1}k) t d^{2} + 2 t^{2} d,
\end{aligned}
\end{equation}
where $\mathcal{F}_{MSA}^{n}$, $\mathcal{F}_{FFN}^{n}$, and $\mathcal{F}_{CFFN}^{n}$ are the FLOPs of the MSA, FFN, and CFFN, respectively. 

Then if we consider $N$ layers of VTE and STE with input sequence length $T$, dimension $D$, strided factor $S$, and kernel size $K$, the encoder-wise FLOPs of VTE $\mathcal{F}_{VTE}$ can be formulated as:
\begin{equation}
   \mathcal{F}_{VTE} =N\mathcal{F}_{VTE}^{n} = {N}(8 T D^{2} + 2T^{2} D),
\end{equation}
the encoder-wise FLOPs of STE $\mathcal{F}_{STE}$ can be formulated as:
\begin{equation}
   \begin{aligned}
      \mathcal{F}_{STE} &=\sum_{n=1}^{N}\mathcal{F}_{STE}^{n} \\[0.5mm] 
      &= \sum_{n=1}^{N}\left[(\frac{6+2KS^{-1}}{S^{n-1}}) T D ^{2} + \frac{2}{S^{2(n-1)}}T^{2} D\right]. 
   \end{aligned}
\end{equation}

For our 27-frame Strided Transformer, which contains $N_{1}$ VTE layers and $N_{2}$ STE layers with $N_{1}=N_{2}=N=3$, $S=3$, and $K=3$.
In this case, the compression ratio $\alpha$ can be computed by: 
\begin{equation}
   \alpha = \frac{2\mathcal{F}_{VTE}}{\mathcal{F}_{VTE}+\mathcal{F}_{STE}} =\frac{2}{1+\beta}, 
\end{equation}
where 
\begin{equation}
   \beta = \frac{\mathcal{F}_{STE}}{\mathcal{F}_{VTE}} = \frac{468D+91T}{972D+243T}.
\end{equation}

We have $\lim _{D \rightarrow \infty} \alpha=1.35$ with a fixed $T$. 
Thus, the compression ratio $\alpha$ of our 27-frame Strided Transformer is 1.35. 

\begin{table*}[t]
   \centering
   \caption
   {
      Quantitative comparisons on Human3.6M under protocol \#1 and protocol \#2,
      where $\dag$ indicates the temporal information used in each method. 
      Best in bold, second-best underlined.
   }
   \resizebox{\textwidth}{!}{
   \begin{tabular}{@{}l|ccccccccccccccc|c@{}}
      \toprule[1pt]
      \textbf{Protocol \#1} & Dir. & Disc & Eat & Greet & Phone & Photo & Pose & Purch. & Sit & SitD. & Smoke & Wait & WalkD. & Walk & WalkT. & Avg.\\
      \midrule[0.5pt]
      
      % A Simple Yet Effective Baseline for 3d Human Pose Estimation
      Martinez \emph{et al.}~\cite{martinez2017simple} ICCV'17  &51.8 &56.2 &58.1 &59.0 &69.5 &78.4 &55.2 &58.1 &74.0 &94.6 &62.3 &59.1 &65.1 &49.5 &  52.4 &62.9\\

      % Learning Pose Grammar to Encode Human Body Configuration for 3D Pose Estimation
      Fang \emph{et al.}~\cite{fang2018learning} AAAI'18  & 50.1& 54.3& 57.0& 57.1& 66.6& 73.3& 53.4& 55.7& 72.8& 88.6& 60.3& 57.7& 62.7& 47.5& 50.6& 60.4 \\

      % Propagating LSTM: 3D Pose Estimation Based on Joint Interdependency
      Lee \emph{et al.}~\cite{lee2018propagating} ECCV'18 $\dag$ &\underline{40.2} &49.2 &47.8 &52.6 &50.1 &75.0 &50.2 &43.0 &55.8 &73.9 &54.1 &55.6 &58.2 &43.3 &43.3 &52.8 \\

      % Graph Stacked Hourglass Networks for 3D Human Pose Estimation
      Xu \emph{et al.}~\cite{xu2021graph} CVPR'21 &45.2 &49.9 &47.5 &50.9 &54.9 &66.1 &48.5 &46.3 &59.7 &71.5 &51.4 &48.6 &53.9 &39.9 &44.1 &51.9 \\

      % PoseAug: A Differentiable Pose Augmentation Framework for 3D Human Pose Estimation
      Gong \emph{et al.}~\cite{gong2021poseaug} CVPR'21  &- &- &- &- &- &- &- &- &- &- &- &- &- &- &- &50.2 \\
      
      % Exploiting Spatial-temporal Relationships for 3D Pose Estimation via Graph Convolutional Networks
      Cai \emph{et al.}~\cite{cai2019exploiting} ICCV'19 $\dag$ &44.6 &47.4 &45.6 &48.8 &50.8 &59.0 &47.2 &43.9&57.9 &61.9 &49.7 &46.6 &51.3 &37.1 &39.4 &48.8 \\
      
      % 3D human pose estimation in video with temporal convolutions and semi-supervised training
      Pavllo \emph{et al.}~\cite{pavllo20193d} CVPR'19 $\dag$ & 45.2 & 46.7 & 43.3 & 45.6 & 48.1 & 55.1 & 44.6 & 44.3 & 57.3 & 65.8 & 47.1 & 44.0 & 49.0 & 32.8 & 33.9 & 46.8 \\

      % Trajectory Space Factorization for Deep Video-Based 3D Human Pose Estimation
      Lin \emph{et al.}~\cite{lin2019trajectory} BMVC'19  $\dag$ &42.5 &44.8 &42.6 &44.2 &48.5 &57.1 &42.6 &41.4 &56.5 &64.5 &47.4 &43.0 &48.1 &33.0 &35.1 &46.6 \\

      % Deep Kinematics Analysis for Monocular 3D Human Pose Estimation
      Xu \emph{et al.}~\cite{xu2020deep} CVPR'20 $\dag$& \textbf{37.4} & {43.5}& 42.7& {42.7}& 46.6& 59.7& \textbf{41.3} &45.1 &\textbf{52.7} &\textbf{60.2} & 45.8& 43.1& 47.7& 33.7& 37.1& 45.6 \\
   
      % Attention Mechanism Exploits Temporal Context: Real-Time 3D Human Pose Reconstruction
      Liu \emph{et al.}~\cite{liu2020attention} CVPR'20  $\dag$ &41.8 &44.8 &{41.1} &44.9 &47.4 &54.1 &43.4 &42.2 &56.2 &63.6 &\underline{45.3} &43.5 &{45.3} &\underline{31.3} &32.2 &45.1 \\

      % Srnet: Improving generalization in 3d human pose estimation with a split-and-recombine approach
      Zeng \emph{et al.}~\cite{zeng2020srnet} ECCV'20  $\dag$& 46.6& 47.1& 43.9& \underline{41.6}& \underline{45.8} & \textbf{49.6} & 46.5& \textbf{40.0}&\underline{53.4} & 61.1& 46.1& 42.6& \textbf{43.1}& 31.5& 32.6& 44.8 \\

      % Motion Guided 3D Pose Estimation from Videos
      Wang \emph{et al.}~\cite{wang2020motion} ECCV'20  $\dag$ &\underline{40.2} &\textbf{42.5} &42.6 &\textbf{41.1} &46.7 &56.7 &\underline{41.4} &42.3 &56.2 &\underline{60.4} &46.3 &\underline{42.2} &46.2 &31.7 &\underline{31.0} &44.5 \\

      % Anatomy-aware 3D Human Pose Estimation with Bone-based Pose Decomposition
      Chen \emph{et al.}~\cite{chen2021anatomy} TCSVT'21  $\dag$ &41.4 &{43.5} &\textbf{40.1} &42.9 &46.6 &\underline{51.9} &41.7 &42.3 &53.9 &\textbf{60.2} &45.4 &\textbf{41.7} &46.0 &31.5 &32.7 &\underline{44.1} \\
      
      \midrule[0.5pt]

      Ours $\dag$ &{40.3} &\underline{43.3} &\underline{40.2} &{42.3} &\textbf{45.6} &{52.3} &{41.8} &\underline{40.5} &{55.9} &{60.6} &\textbf{44.2} &{43.0} &\underline{44.2} &\textbf{30.0} &\textbf{30.2} &\textbf{43.7} \\

      \toprule[1pt]

      \textbf{Protocol \#2} & Dir. & Disc & Eat & Greet & Phone & Photo & Pose & Purch. & Sit & SitD. & Smoke & Wait & WalkD. & Walk & WalkT. & Avg.\\
      \midrule[0.5pt]

      % A Simple Yet Effective Baseline for 3d Human Pose Estimation
      Martinez \emph{et al.}~\cite{martinez2017simple} ICCV'17  &39.5 &43.2 &46.4 &47.0 &51.0 &56.0 &41.4 &40.6 &56.5 &69.4 &49.2 &45.0 &49.5 &38.0 &43.1 &47.7 \\

      % Ordinal Depth Supervision for 3D Human Pose Estimation
      Pavlakos \emph{et al.}~\cite{pavlakos2018ordinal} CVPR'18  &34.7 &39.8 &41.8 &38.6 &42.5 &47.5 &38.0 &36.6 &50.7 &56.8 &42.6 &39.6 &43.9 &32.1 &36.5 &41.8 \\

      % A comprehensive study of weight sharing in graph networks for 3d human pose estimation
      Liu \emph{et al.}~\cite{liu2020comprehensive} ECCV'20  &35.9 &40.0 &38.0 &41.5 &42.5 &51.4 &37.8 &36.0 &48.6 &56.6 &41.8 &38.3 &42.7 &31.7 &36.2 &41.2 \\

      % PoseAug: A Differentiable Pose Augmentation Framework for 3D Human Pose Estimation
      Gong \emph{et al.}~\cite{gong2021poseaug} CVPR'21  &- &- &- &- &- &- &- &- &- &- &- &- &- &- &-  &39.1 \\

      % Exploiting Spatial-temporal Relationships for 3D Pose Estimation via Graph Convolutional Networks
      Cai \emph{et al.}~\cite{cai2019exploiting} ICCV'19  $\dag$ &35.7 &37.8 &36.9 &40.7 &39.6 &45.2 &37.4 &34.5 &46.9 &50.1 &40.5 &36.1 &41.0 &29.6 &33.2 &39.0 \\

      % Trajectory Space Factorization for Deep Video-Based 3D Human Pose Estimation
      Lin \emph{et al.}~\cite{lin2019trajectory} BMVC'19  $\dag$ &32.5 &35.3 &34.3 &36.2 &37.8 &43.0 &33.0 &\underline{32.2} &45.7 &51.8 &38.4 &\underline{32.8} &37.5 &25.8 &28.9 &36.8 \\

      % 3D human pose estimation in video with temporal convolutions and semi-supervised training
      Pavllo \emph{et al.}~\cite{pavllo20193d} CVPR'19 $\dag$ &34.1 &36.1 &{34.4} &37.2 &{36.4} &42.2 &34.4 &33.6 &45.0 &52.5 &37.4 &33.8 &37.8 &25.6 &27.3 &36.5\\

      % Deep Kinematics Analysis for Monocular 3D Human Pose Estimation
      Xu \emph{et al.}~\cite{xu2020deep} CVPR'20 $\dag$ &\textbf{31.0} &\textbf{34.8} &34.7 &\textbf{34.4} &\underline{36.2} &43.9 &\underline{31.6} &33.5 &\textbf{42.3} &\textbf{49.0} &37.1 &33.0 &39.1 &26.9 &31.9 &36.2 \\

      % Attention Mechanism Exploits Temporal Context: Real-Time 3D Human Pose Reconstruction
      Liu~et al.~\cite{liu2020attention} CVPR'20 $\dag$ &\underline{32.3} &\underline{35.2} &\underline{33.3} &35.8 &\textbf{35.9} &\textbf{41.5} &33.2 &{32.7} &\underline{44.6} &50.9 &\underline{37.0} &\textbf{32.4} &37.0 &\underline{25.2} &27.2 &35.6 \\

      % Motion Guided 3D Pose Estimation from Videos
      Wang \emph{et al.}~\cite{wang2020motion} ECCV'20 $\dag$  &32.9 &\underline{35.2} &35.6 &\textbf{34.4} &36.4 &42.7 &\textbf{31.2} &{32.5} &45.6 &50.2 &37.3 &\underline{32.8} &\underline{36.3} &26.0 &\textbf{23.9} &\underline{35.5} \\

      \midrule[0.5pt]

      Ours $\dag$ &{32.7} &{35.5} &\textbf{32.5} &\underline{35.4} &\textbf{35.9} &\underline{41.6} &{33.0} &\textbf{31.9} &{45.1} &\underline{50.1} &\textbf{36.3} &{33.5} &\textbf{35.1} &\textbf{23.9} &\underline{25.0}  &\textbf{35.2} \\

      \toprule[1pt]
      \end{tabular}
   }
   \label{table:h36m}
\end{table*}

\begin{table}[ht]
   \centering
   \caption
   { 
      Quantitative comparisons with state-of-the-art methods in different receptive fields on Human3.6M. 
      The computational complexity, MPJPE, and frame per second (FPS) are reported. 
      FPS is computed on a single GeForce GTX 2080 Ti GPU.
   }  
   \setlength{\tabcolsep}{1.10mm} 
   \begin{tabular}{cccccc}
   \toprule [1pt]
   Model &$T$ &Param (M) &FLOPs (G) &MPJPE (mm) &FPS \\
   \midrule [0.5pt]  
   Pavllo \emph{et al.}~\cite{pavllo20193d} &27 &8.56 &0.017 &48.8 &1492 \\
   Pavllo \emph{et al.}~\cite{pavllo20193d} &81 &12.75 &0.025 &47.7 &1121 \\
   Pavllo \emph{et al.}~\cite{pavllo20193d} &243 &16.95 &0.033 &46.8 &863 \\
   Chen \emph{et al.}~\cite{chen2021anatomy} &27 &31.88 &0.061 &45.3 &410 \\
   Chen \emph{et al.}~\cite{chen2021anatomy} &81 &45.53 &0.088  &44.6 &315 \\
   Chen \emph{et al.}~\cite{chen2021anatomy} &243 &59.18 &0.116 &44.1 &264 \\
   \midrule [0.5pt]

   Ours (27 frames) &27 &4.01 &0.128 &46.9 &118 \\
   Ours (81 frames) &81 &4.06 &0.392 &45.4 &112 \\
   Ours (243 frames) &243 &4.23 &1.372 &44.0 &108 \\
   Ours (351 frames) &351 &4.34 &2.142 &\textbf{43.7} &105 \\

   \toprule [1pt]
   \end{tabular}
   \label{table:compare}
\end{table}

\section{Experiments}
\subsection{Datasets and Evaluation}
The proposed method is evaluated on two challenging benchmark datasets, \emph{i.e.}, Human3.6M~\cite{ionescu2013human3} and HumanEva-I~\cite{sigal2010humaneva}. 
Human3.6M dataset is the largest publicly available dataset for 3D human pose estimation, which consists of 3.6 million images captured from 4 synchronized cameras with 50 Hz. 
There are 7 professional subjects performing 15 daily activities such as “Waiting”, “Smoking”, and “Posing”.
Following the standard protocol in prior works~\cite{chen2019weakly,tome2018rethinking,pavllo20193d}, 5 subjects (S1, S5, S6, S7, S8) are used for training and 2 subjects (S9 and S11) are used for evaluation. 
The frames from all views are trained by a single model for all actions. 
HumanEva-I is a much smaller dataset with fewer subjects and actions compared to Human3.6M. 
Following~\cite{pavllo20193d,lee2018propagating}, our model is trained for all subjects (S1, S2, S3) and all actions (Walk, Jog, Box). 

\begin{table*}[htb]
   \centering
   \caption
   {
      Quantitative comparisons of MPJPE in millimeter on Human3.6M under protocol \#1, using ground truth 2D joint locations as input. 
      $\dag$ means the method utilizing temporal information. 
      Best in bold. 
   }
   \resizebox{\textwidth}{!}{
   \begin{tabular}{@{}l|ccccccccccccccc|c@{}}
   \toprule[1pt]
   \textbf{Protocol \#1} & Dir. & Disc & Eat & Greet & Phone & Photo & Pose & Purch. & Sit & SitD. & Smoke & Wait & WalkD. & Walk & WalkT. & Avg.\\
   \midrule[0.5pt]

   % A Simple Yet Effective Baseline for 3d Human Pose Estimation
   Martinez \emph{et al.}~\cite{martinez2017simple} ICCV'17 &37.7 &44.4 &40.3 &42.1 &48.2 &54.9 &44.4 &42.1 &54.6 &58.0 &45.1 &46.4 &47.6 &36.4 &40.4 &45.5 \\

   % Propagating LSTM: 3D Pose Estimation Based on Joint Interdependency
   Lee \emph{et al.}~\cite{lee2018propagating} ECCV'18 $\dag$  &32.1 &36.6 &34.3 &37.8 &44.5 &49.9 &40.9 &36.2 &44.1 &45.6 &35.3 &35.9 &30.3 &37.6 &35.5 &38.4 \\

   % 3D human pose estimation in video with temporal convolutions and semi-supervised training
   Pavllo \emph{et al.}~\cite{pavllo20193d} CVPR'19 $\dag$  &35.2 &40.2 &32.7 &35.7 &38.2 &45.5 &40.6 &36.1 &48.8 &47.3 &37.8 &39.7 &38.7 &27.8 & 29.5 &37.8 \\

   % Exploiting Spatial-temporal Relationships for 3D Pose Estimation via Graph Convolutional Networks
   Cai \emph{et al.}~\cite{cai2019exploiting} ICCV'19 $\dag$  &32.9 &38.7 &32.9 &37.0 &37.3 &44.8 &38.7 &36.1 &41.0 &45.6 &36.8 &37.7 &37.7 &29.5 &31.6 &37.2 \\

   % Graph Stacked Hourglass Networks for 3D Human Pose Estimation
   Xu \emph{et al.}~\cite{xu2021graph} CVPR'21 &35.8 &38.1 &31.0 &35.3 &35.8 &43.2 &37.3 &31.7 &38.4 &45.5 &35.4 &36.7 &36.8 &27.9 &30.7 &35.8 \\

   % Attention Mechanism Exploits Temporal Context: Real-Time 3D Human Pose Reconstruction
   Liu \emph{et al.}~\cite{liu2020attention} CVPR'20 $\dag$ &34.5 &37.1 &33.6 &34.2 &32.9 &37.1 &39.6 &35.8 &40.7 &41.4 &33.0 &33.8 &33.0 &26.6 &26.9 &34.7 \\

   %  Anatomy-aware 3D Human Pose Estimation with Bone-based Pose Decomposition
   Chen \emph{et al.}~\cite{chen2021anatomy} TCSVT'21 $\dag$  &- &- &- &- &- &- &- &- &- &- &- &- &- &- &- &32.3 \\
   
   % Srnet: Improving generalization in 3d human pose estimation with a split-and-recombine approach
   Zeng \emph{et al.}~\cite{zeng2020srnet} ECCV'20 $\dag$  &34.8 &32.1 &28.5 &30.7 &31.4 &36.9 &35.6 &30.5 &38.9 &40.5 &32.5 &31.0 &29.9 &22.5 &24.5 &32.0 \\

   \midrule[0.5pt]
   Ours &\textbf{27.1} &\textbf{29.4} &\textbf{26.5} &\textbf{27.1} &\textbf{28.6} &\textbf{33.0} &\textbf{30.7} &\textbf{26.8} &\textbf{38.2} &\textbf{34.7} &\textbf{29.1} &\textbf{29.8} &\textbf{26.8} &\textbf{19.1} &\textbf{19.8} &\textbf{28.5} \\

   \toprule[1pt]
   \end{tabular}
   }
   \label{table:h36m_gt}
\end{table*}

\begin{table*}[htb]
   \centering
   \caption
   {
      Results show the velocity error (MPJPV) of our methods and other state-of-the-arts on Human3.6M. 
      Here, $^{*}$ denotes our result without the supervision of full sequence scale. 
      Best in bold. 
   }
   \resizebox{\textwidth}{!}{
   \begin{tabular}{@{}l|ccccccccccccccc|c@{}}
   \toprule[1pt]
   \textbf{MPJPV} & Dir. & Disc & Eat & Greet & Phone & Photo & Pose & Purch. & Sit & SitD. & Smoke & Wait & WalkD. & Walk & WalkT. & Avg.\\
   \midrule[0.5pt]
   % 3D human pose estimation in video with temporal convolutions and semi-supervised training
   Pavllo \emph{et al.}~\cite{pavllo20193d} CVPR'19   &3.0 &3.1 &2.2 &3.4 &2.3 &2.7 &2.7 &3.1 &2.1 &2.9 &2.3 &2.4 &3.7 &3.1 &2.8 &2.8 \\

   % Trajectory Space Factorization for Deep Video-Based 3D Human Pose Estimation
   Lin \emph{et al.}~\cite{lin2019trajectory} BMVC'19   &2.7 &2.8 &2.1 &3.1 &2.0 &2.5 &2.5 &2.9 &1.8 &2.6 &2.1 &2.3 &3.7 &2.7 &3.1 &2.7 \\

   %  Anatomy-aware 3D Human Pose Estimation with Bone-based Pose Decomposition
   Chen \emph{et al.}~\cite{chen2021anatomy} TCSVT'21 &2.7 &2.8 &2.0 &3.1 &2.0 &2.4 &2.4 &2.8 &1.8 &2.4 &2.0 &2.1 &3.4 &2.7 &2.4 &2.5 \\

   % Motion Guided 3D Pose Estimation from Videos
   Wang \emph{et al.}~\cite{wang2020motion} ECCV'20 &\textbf{2.3} &\textbf{2.5} &2.0 &\textbf{2.7} &2.0 &2.3 &\textbf{2.2} &\textbf{2.5} &1.8 &2.7 &1.9 &2.0 &\textbf{3.1} &\textbf{2.2} &2.5 &2.3 \\
   
   \midrule[0.5pt]
   Ours $^{*}$ &2.8 &2.8 &2.1 &3.2 &2.2 &2.5 &2.6 &2.8 &1.8 &2.4 &2.1 &2.3 &3.5 &3.0 &2.6 &2.6 \\

   Ours &{2.4} &\textbf{2.5} &\textbf{1.8} &{2.8} &\textbf{1.8} &\textbf{2.2} &\textbf{2.2} &\textbf{2.5} &\textbf{1.5} &\textbf{2.0} &\textbf{1.8} &\textbf{1.9} &{3.2} &{2.5} &\textbf{2.1} &\textbf{2.2} \\

   \toprule[1pt]
   \end{tabular}
   }
   \label{table:mpjpv}
\end{table*}

\begin{table}[ht]
   \centering
   \scriptsize
   \caption
   {
      Quantitative results on HumanEva-I dataset under protocol \#2. 
      Best in bold, second-best underlined.
      (MRCNN) - Mask-RCNN; 
      (GT) - 2D ground truth. 
   }
   \setlength{\tabcolsep}{0.95mm} 
   \begin{tabular}{@{}l|ccc|ccc|ccc|c@{}}
   \toprule[1pt]
   & \multicolumn{3}{c}{Walk} & \multicolumn{3}{c}{Jog} &
   \multicolumn{3}{c}{Box} \\
   & S1 & S2 & S3 & S1 & S2 & S3 & S1 & S2 & S3 & Avg. \\
   \midrule[0.5pt]

   Martinez \emph{et al.}~\cite{martinez2017simple} &19.7 &17.4 &46.8 &26.9 &18.2 &18.6 &- &- &- &-\\

   Pavlakos \emph{et al.}~\cite{pavlakos2017coarse} &22.3 &19.5 &\textbf{29.7} &28.9 &21.9 &23.8 &- &- &- &-\\

   Lee \emph{et al.}~\cite{lee2018propagating} &18.6 &19.9 &\underline{30.5} &25.7 &16.8 &17.7 &42.8 &48.1 &53.4 &30.3 \\

   Pavllo \emph{et al.}~\cite{pavllo20193d} &\textbf{13.9} &\underline{10.2} &46.6 &\underline{20.9} &\textbf{13.1} &\textbf{13.8} &\underline{23.8} &\underline{33.7} &\underline{32.0} &\underline{23.1} \\

   \midrule[0.5pt]
   Ours ($T=27$ MRCNN) &\underline{14.0} &\textbf{10.0} &32.8 &\textbf{19.5} &\underline{13.6} &\underline{14.2} &\textbf{22.4} &\textbf{21.6} &\textbf{22.5} &\textbf{18.9} \\
   \midrule[0.5pt]

   Ours ($T=27$ GT) &9.7 &7.6 &15.8 &12.3 &9.4 &11.2 &14.8 &12.9 &16.5 &12.2 \\
   \toprule[1pt]
   \end{tabular}
   \label{table:humaneva_eval}
\end{table} 

Three standard evaluation protocols are used in the experiments. 
The mean per joint position error (MPJPE) is the average Euclidean distance between the ground truth and predicted positions of the joints, which is referred to as protocol \#1 in many works~\cite{fang2018learning,kocabas2019self}. 
Procrustes analysis MPJPE (P-MPJPE) is adopted, where the estimated 3D pose is aligned to the ground truth in translation, rotation, and scale. 
This protocol is referred to as protocol \#2~\cite{martinez2017simple,rayat2018exploiting}.
Following~\cite{pavllo20193d,chen2021anatomy,wang2020motion}, we also report the mean per joint velocity error (MPJVE) corresponding to the MPJPE of the first derivative of the 3D pose sequences. 
This metric measures the smoothness of predictions over time and is vital for video-based 3D pose estimation.

\subsection{Implementation Details}
In our experiments, the proposed Strided Transformer contains $N_{1}=N_{2}=3$ encoder layers, $h=8$ attention heads, $d_m=256$ dimensions, and $d_f=512$ hidden units for both VTE and STE. 
The kernel sizes $k_{f}$ and $k_{m}$ are set to 1 and 3 in all STE layers, respectively. 
The strided factor $s_f$ is set to 1, and $s_m$ is set to $\left\{3, 3, 3 \right\}$ for the receptive field of 27 frames, $\left\{9, 3, 3\right\}$ for 81, $\left\{3, 9, 9\right\}$ for 243, and $\left\{3, 9, 13\right\}$ for 351. 
The weighting factors $\lambda_{f}$ and $\lambda_{s}$ are set to 1. 

All experiments are conducted on the PyTorch framework with one GeForce GTX 3090 GPU.
The network is trained using Amsgrad optimizer. 
An initial learning rate of 0.001 is used with a shrink factor of 0.95  applied after each epoch. 
The same refine module as~\cite{cai2019exploiting,wang2020motion} is adopted. 
We only apply horizontal flip augmentation during training/test stages. 
The 2D poses can be obtained by performing any classic 2D pose detections or directly using the 2D ground truth. 
Following~\cite{pavllo20193d,cheng2019occlusion}, the cascaded pyramid network (CPN)~\cite{chen2018cascaded} is used for Human3.6M and Mask R-CNN~\cite{he2017mask} is adopted for HumanEva-I to obtain 2D poses for a fair comparison.

\subsection{Comparison with State-of-the-art Results}
Our method is compared with previous state-of-the-art approaches on Human3.6M dataset. 
The performance of our 351-frame model with CPN input is reported in Table~\ref{table:h36m}. 
Our method outperforms the state-of-the-art methods on Human3.6M under all metrics (43.7 mm on protocol \#1 and 35.2 mm on protocol \#2). 

\begin{figure*}[htb]
   \centering
   \includegraphics[width=1.00 \linewidth]{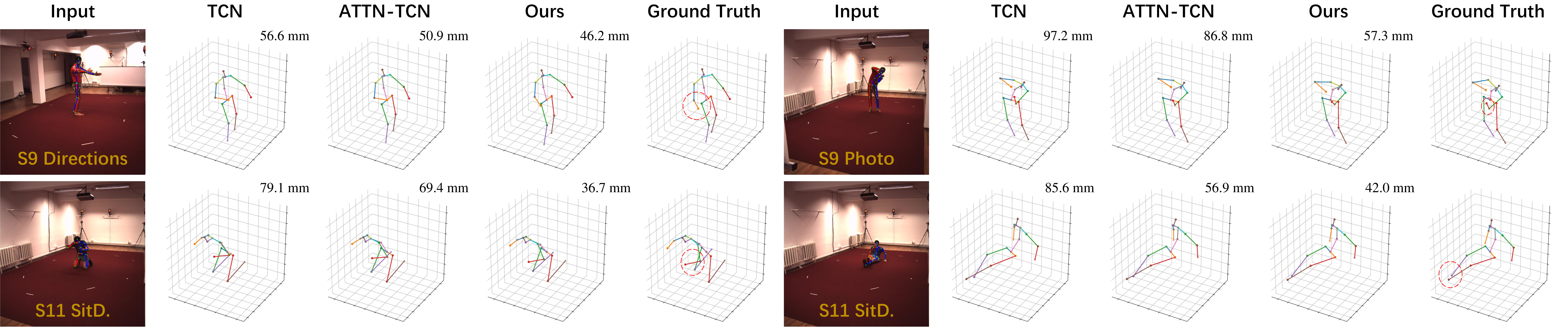}
   \caption
   {
      Qualitative comparisons with the previous state-of-the-art methods, TCN~\cite{pavllo20193d} and ATTN-TCN~\cite{liu2020attention} on Human3.6M dataset. 
      Wrong estimations are highlighted by red circles. 
   }
   \label{fig:results}
\end{figure*}

Table~\ref{table:compare} compares the computational complexity, MPJPE, and frame per second (FPS) with several state-of-the-art methods in different receptive fields on Human3.6M. 
Our model is lightweight and the number of parameters hardly increases with the increased receptive fields, which is practical for real-time applications. 
Compared with temporal convolutional networks~\cite{pavllo20193d,chen2021anatomy}, our proposed Transformer-based network requires fewer total parameters with competitive performance for 3D pose estimation in videos. 
Besides, even though the inference speed of the proposed model is lower than~\cite{pavllo20193d,chen2021anatomy}, it still has an acceptable FPS for real-time inference. 
Fig.~\ref{fig:results} shows some qualitative comparisons with state-of-the-art methods~\cite{pavllo20193d,liu2020attention}, which indicates that our methods can produce more accurate 3D predictions. 

To further explore the upper bound of our method, the results from 2D ground truth inputs are reported in Table~\ref{table:h36m_gt}. 
It can be seen that our method achieves the best result (28.5 mm in MPJPE), outperforming all other methods. 
This demonstrates if a more robust 2D pose detection is available, our Strided Transformer can produce more accurate 3D poses. 

As shown in Table~\ref{table:mpjpv}, with the supervision of full sequence scale, our method reduces the MPJVE by 15.4\% (from 2.6 mm to 2.2 mm), achieving smoother predictions with lower MPJVE than other models.
It indicates that the full-to-single supervision scheme can enhance temporal smoothness and produce vastly smoother poses. 

To evaluate the generalizability of our model to smaller datasets, experiments are conducted on HumanEva-I based on Mask R-CNN 2D detections and 2D ground truth. 
The results in Table~\ref{table:humaneva_eval} demonstrate that our method achieves promising results on all kinds of actions. 

\subsection{Ablation Studies}
\begin{figure}[tb]
   \centering
   \begin{subfigure}[htb]{0.241\textwidth}
      \includegraphics[width=\textwidth]{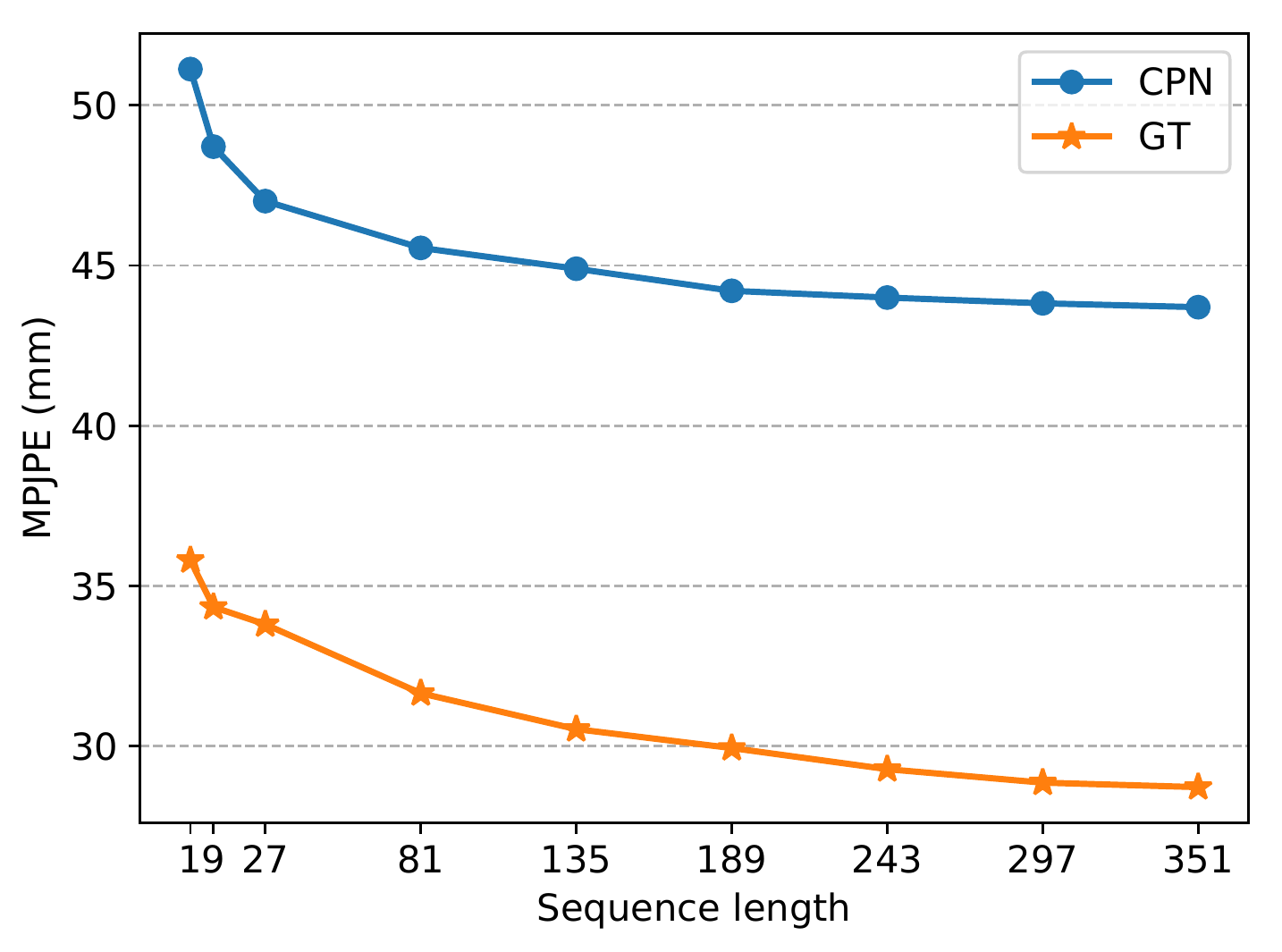}
      \caption{}
      \label{fig:frames}
   \end{subfigure}
   \begin{subfigure}[htb]{0.241\textwidth}
      \includegraphics[width=\textwidth]{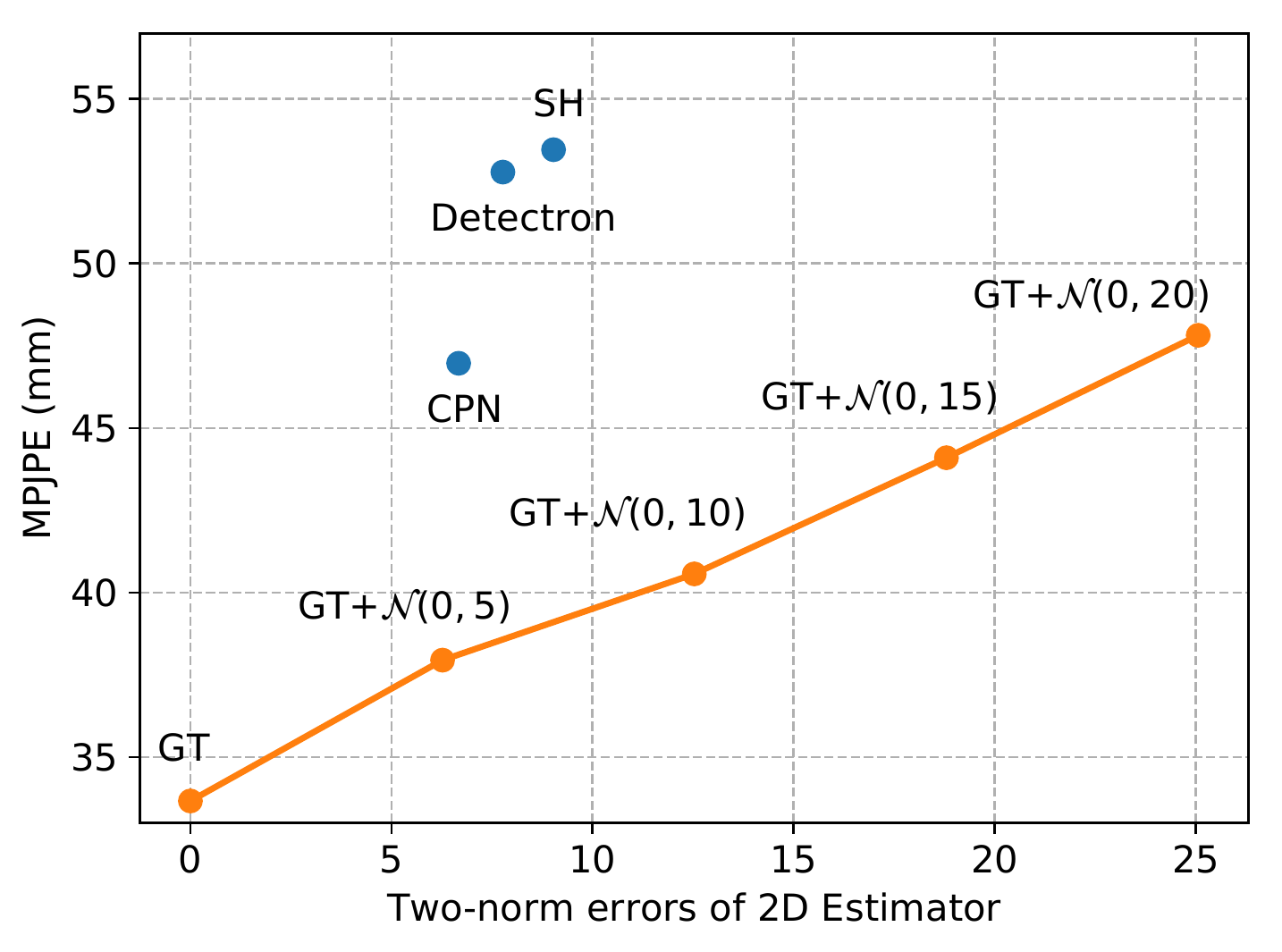}
      \caption{}
      \label{fig:2D_detections}
   \end{subfigure}
   \caption
   {
      (a) Ablation studies on different sequence lengths of our method on Human3.6M with the MPJPE metric. 
      (b) The impact of 2D detections on Human3.6M. 
      Here, $\mathcal{N}(0, \sigma^2)$ represents the Gaussian noise with mean zero and $\sigma$ is the standard deviation. 
      (CPN) - Cascaded Pyramid Network; (SH) Stack Hourglass; (GT) - 2D ground truth.  
   }
   \label{fig:frames and 2D}
\end{figure}

\textbf{Input sequence length.}
The MPJPE results of our model with different sequence lengths (between 1 and 351) on Human3.6M are shown in Fig.~\ref{fig:frames and 2D} (a). 
It can be seen that our proposed method obtains larger gains under both 2D pose inputs (CPN and GT) with more input frames used for predictions, but the error saturates past a certain point. 
This is expected since directly lifting 3D poses from disjointed 2D poses leads to temporally incoherent outputs~\cite{dabral2018learning}. 
It is worth mentioning that our method gets a better result with $T=351$ (43.7 mm) than $T=243$ (44.0 mm), while the performance decreases with longer inputs ($T>243$) in~\cite{liu2020attention}. 
This indicates that our method equipped with the global self-attention mechanism is powerful in modeling long-range dependencies. 
Meanwhile, with the help of STE, our method can learn the representative representation from long sequences. 
Next, we choose $T = 27$ on Human3.6M in the following ablation experiments as a compromise between the accuracy and computational complexity. 

\textbf{2D detections.}
For the 2D-to-3D pose lifting task, the accuracy of the 2D detections directly influences the results of 3D pose estimation~\cite{martinez2017simple}. 
To show the effectiveness of our method on different 2D pose detectors, we carry out experiments with the detections from Stack Hourglass (SH)~\cite{newell2016stacked}, Detectron~\cite{pavllo20193d}, and CPN~\cite{chen2018cascaded}. 
Moreover, to test the tolerance of our method to different levels of noise, we also train our network by 2D ground truth (GT) with various levels of additive Gaussian noises. 
The results are shown in Fig.~\ref{fig:frames and 2D} (b). 
It can be observed that the MPJPE of 3D poses increases linearly with the two-norm errors of 2D poses. 
Besides, our method performs well on different 2D inputs, indicating the effectiveness and robustness of our method.

\textbf{Model hyperparameters.}
As shown in Table~\ref{table:hyperparameters}, we first analyze the effect of the number of VTE layers. 
Empirically, it can be found that the performance cannot be improved when naively stacking multiple standard Transformer encoder layers. 
Notably, our model equipped with STE is more accurate at the same number of Transformer encoder layers and comparable model parameters. 
For example, our method ($N_{1}=3$ and $N_{2}=3$) has better performance and fewer FLOPs than the model of $N_{1}=6$ at the same $d_{m} = 256$ and $d_{f} = 512$ (46.9 mm vs. 47.9 mm, 0.128G vs. 0.174G). 
In addition, our STE ($N_{2}=3$, 0.041G) also has fewer FLOPs than standard Transformer encoder ($N_{1}=3$, 0.087G) with similar parameters, which achieves $2.1\times$ less computation. 
It verifies the effectiveness of our proposed STE in reducing computation cost and boosting performance. 
Then, we investigate the influence of various hyperparameters combinations to find the optimal network architecture. 
It can be observed that using 3 encoder layers of both VTE and STE modules, 256 dimensions, and 512 hidden units achieves the best performance. 

\textbf{Strided factor.}
We observe that the strided factor of STE used in our Strided Transformer has an impact on the estimation performance. 
Here, we study the influence of using different design choices of strided factor of STE. 
The experimental results are depicted in Table~\ref{table:Strided}.
It shows that using a strided factor $s_m = \left\{3, 3, 3\right\}$ has the best performance. 
This demonstrates the benefit of gradually reducing the temporal dimensionality with a small strided factor. 

\begin{table}[tb]
   \centering
   \caption
   {
      Ablation study on the hyperparameters of our model on Human3.6M under protocol \#1. 
      $N_{1}$ and $N_{2}$ are the number of VTE and STE layers, respectively. 
      $d_{m}$ and $d_{f}$ are the dimensions and the number of hidden units. 
   }
   \setlength{\tabcolsep}{1.8mm} 
   \begin{tabular}{ccccccc}
   \toprule  [1pt]
   $N_{1}$ &$N_{2}$ &$d_{m}$ &$d_{f}$ &Param (M) &FLOPs (G)  &MPJPE (mm)\\
   \midrule  [0.5pt]
   2 &- &512 &2048 &6.36 &0.342  &47.9 \\
   3 &- &512 &2048 &9.51 &0.514  &47.8 \\
   4 &- &512 &2048 &12.66 &0.685 &48.0 \\
   5 &- &512 &2048 &15.82 &0.856 &48.4 \\
   6 &- &512 &2048 &18.97 &1.028 &49.3 \\
   \midrule [0.5pt]  

   2 &- &256 &512 &1.08 &0.058 &47.8 \\
   3 &- &256 &512 &1.61 &0.087 &47.6 \\
   4 &- &256 &512 &2.13 &0.116 &47.8 \\
   5 &- &256 &512 &2.66 &0.145 &47.7 \\
   6 &- &256 &512 &3.19 &0.174 &47.9 \\
   \midrule [0.5pt]  

   - &3 &256 &512 &2.42 &0.041 &48.0 \\
   2 &3 &256 &512 &3.48 &0.099 &47.4 \\
   3 &3 &256 &512 &4.01 &0.128 &\textbf{46.9} \\
   2 &3 &512 &2048 &22.18 &0.589 &47.4 \\
   3 &3 &512 &2048 &25.33 &0.761 &47.3 \\
   \toprule  [1.0pt] 
   \end{tabular}
\label{table:hyperparameters}
\end{table}

\begin{table}[tb]
   \caption
   {
      Ablation study on the strided factor of STE with the receptive field $T = 3 \times 3 \times 3 = 27$. 
      The evaluation is performed on Human3.6M under protocol \#1.
   }
   \centering  
   \setlength{\tabcolsep}{7.50mm} 
   \begin{tabular}{ccc}
   \toprule  [1pt]
   Layers &Strided factor &MPJPE (mm)  \\
   \midrule  [0.5pt]  
   3 &$3, 3, 3$ &\textbf{46.9} \\
   3 &$3, 9, 1$ &47.5  \\
   3 &$9, 3, 1$ &47.3 \\
   2 &$3, 9$ &47.2  \\
   2 &$9, 3$ &47.1  \\
   1 &$27$ &47.7 \\
   \toprule [1pt]
   \end{tabular}
   \label{table:Strided}
\end{table}

\begin{table}[tb]
   \caption
   {
      Ablation study on different prediction schemes. 
      The evaluation is performed on Human3.6M under protocol \#1.
      $\Delta$ represents the performance gap between the methods and ours.
   }
   \centering  
   \setlength{\tabcolsep}{7.30mm} 
   \begin{tabular}{lcc}
   \toprule  [1pt]
   Prediction scheme &MPJPE (mm) &$\Delta$ \\
   \midrule [0.5pt]
   Full &47.9 &1.0 \\
   Single &48.3 &1.4 \\
   Full-to-full &47.4 &0.5 \\
   Single-to-single &48.5 &1.6 \\
   Full-to-single &\textbf{46.9} &- \\
   \toprule [1pt]
   \end{tabular}
   \label{table:prediction}
\end{table}

\begin{table}[!htb]
   \centering
   \caption
   {
      Ablation study on each component of our network architecture on Human3.6M under protocol \#1.
   }
   \setlength{\tabcolsep}{8.35mm} 
   
   \begin{tabular}{lc}
   \toprule [1.0pt] 
   Method& MPJPE (mm) \\
   \midrule [0.5pt] 
   Ours, proposed &\textbf{46.9} \\
   Ours, intermediate predictions &48.1 \\
   Ours, Pooling Transformer &47.3 \\
   \midrule [0.5pt]

   w/o VTE &48.0 \\
   w/o STE &47.6 \\
   \toprule [1.0pt] 
   \end{tabular}
   \label{table:ablation_method}
\end{table}

\textbf{Prediction scheme.}
We further examine the proposed prediction scheme of full sequence scale and single target frame scale by using five different designs: 
(i) Full: the STE of our proposed method is replaced with VTE, and the new architecture is only supervised by the full sequence scale (the sequence loss). 
(ii) Single: the proposed method is only supervised by the single target frame scale (single-frame loss). 
(iii) Full-to-full: the architecture consists of six VTE layers, whose first three layers and final three layers are both supervised by the sequence loss. 
(iv) Single-to-single: VTE and STE of the proposed method are both supervised by the single-frame loss.
(v) Full-to-single: our proposed method. 
In Table~\ref{table:prediction}, it can be observed that the schemes of considering only one prediction manner (i, ii, iii, iv) decay performance, and our full-to-single prediction scheme (v) is the best. 
The empirical results indicate that our proposed full-to-single mechanism is crucial for performance improvement. 

\textbf{Model components.}
As shown in Table~\ref{table:ablation_method}, an ablation study is performed to assess the effectiveness of different components of our method. 
We select the center frame of intermediate predictions from VTE as final results, which increases the 
MPJPE by 1.2 mm (from 46.9 mm to 48.1 mm). 
It proves that the scheme of intermediate supervision can further improve estimation accuracy. 
Next, we perform pooling operation after FFN of VTE following~\cite{zihang2020funnel-transformer} and then replace STE of our proposed method with it. 
The new architecture is termed as Pooling Transformer, and its error increases by 0.4 mm, which highlights that our STE can preserve more valuable information than Pooling Transformer by exploiting local contexts to aggregate information. 
Removing VTE (only trained with single-frame loss) leads to a 1.1 mm increase in MPJPE error. 
Besides, removing STE (only trained with sequence loss) increases the MPJPE to 47.6 mm. 
These results validate the importance of both VTE and STE modules in our Strided Transformer, where VTE mainly models long-range information and STE focuses on aggregating information in a hierarchical global and local fashion. 

\begin{figure*}[htb]
   \centering
   \begin{subfigure}[htb]{0.497\textwidth}
      \includegraphics[width=\textwidth]{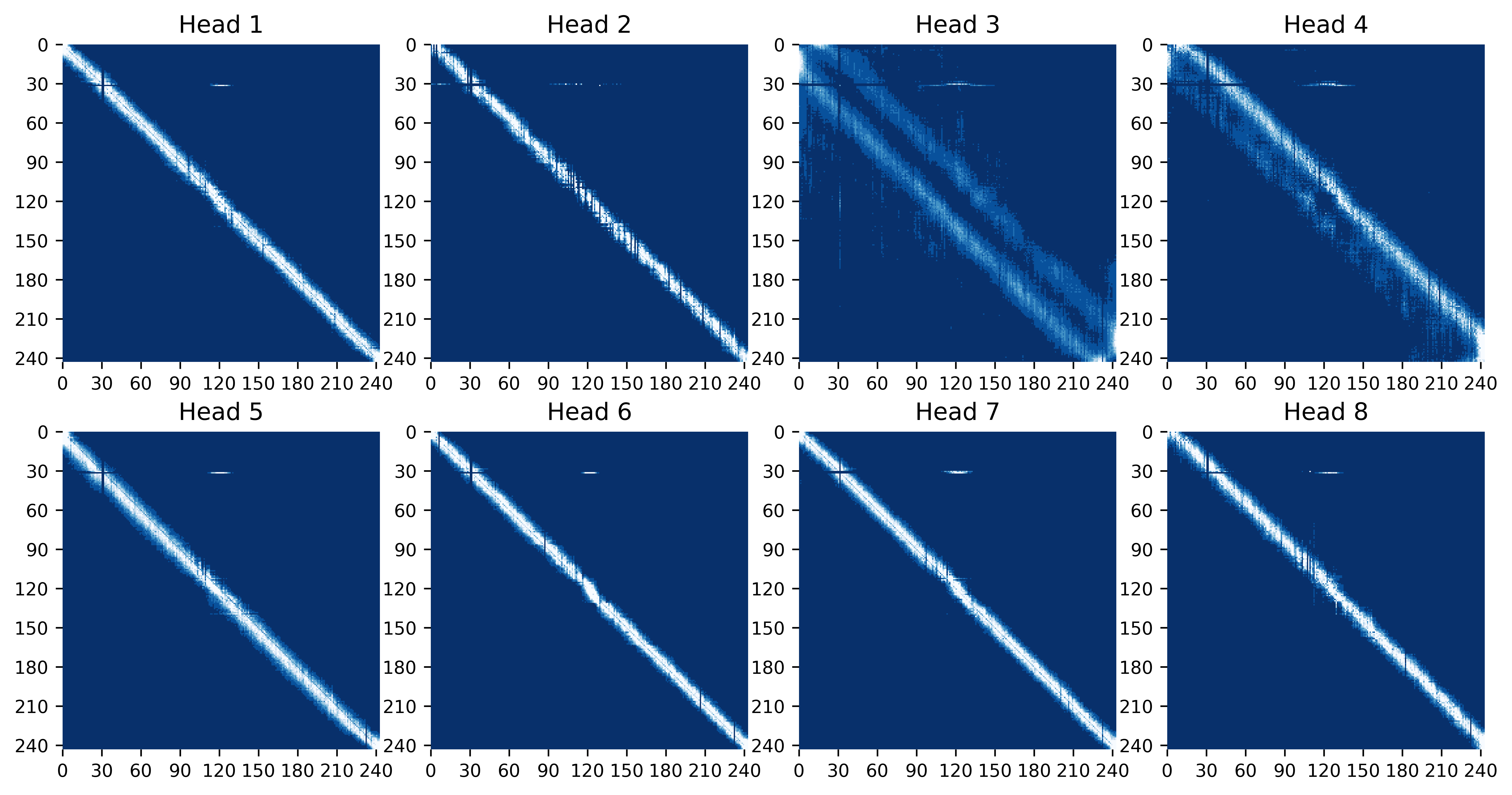}
      \caption{VTE}
      \label{fig:attetion_VTE}
   \end{subfigure}
   \begin{subfigure}[htb]{0.497\textwidth}
      \includegraphics[width=\textwidth]{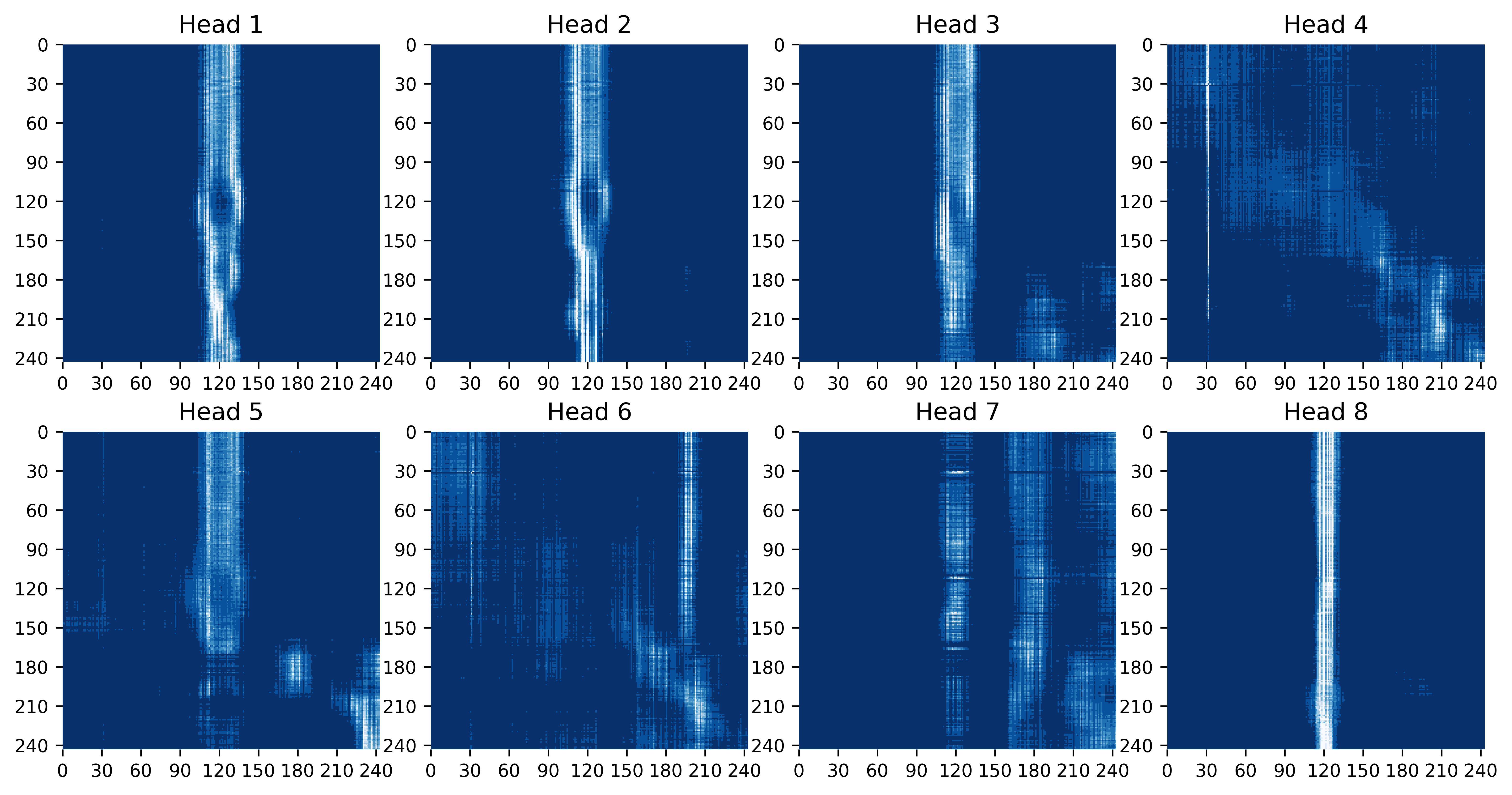}
      \caption{STE}
      \label{fig:attetion_STE}
   \end{subfigure}
   \caption
   {
      Multi-head attention maps ($h=8$) from VTE and STE of our 243-frame model. 
      It illustrates that the self-attention mechanism systematically assigns a weight distribution to frames, all of which might benefit the inference. 
      Brighter color indicates higher attention score. 
   }
   \label{fig:attention}
\end{figure*}

\begin{figure}[htb]
   \centering
   \includegraphics[width=1.0 \linewidth]{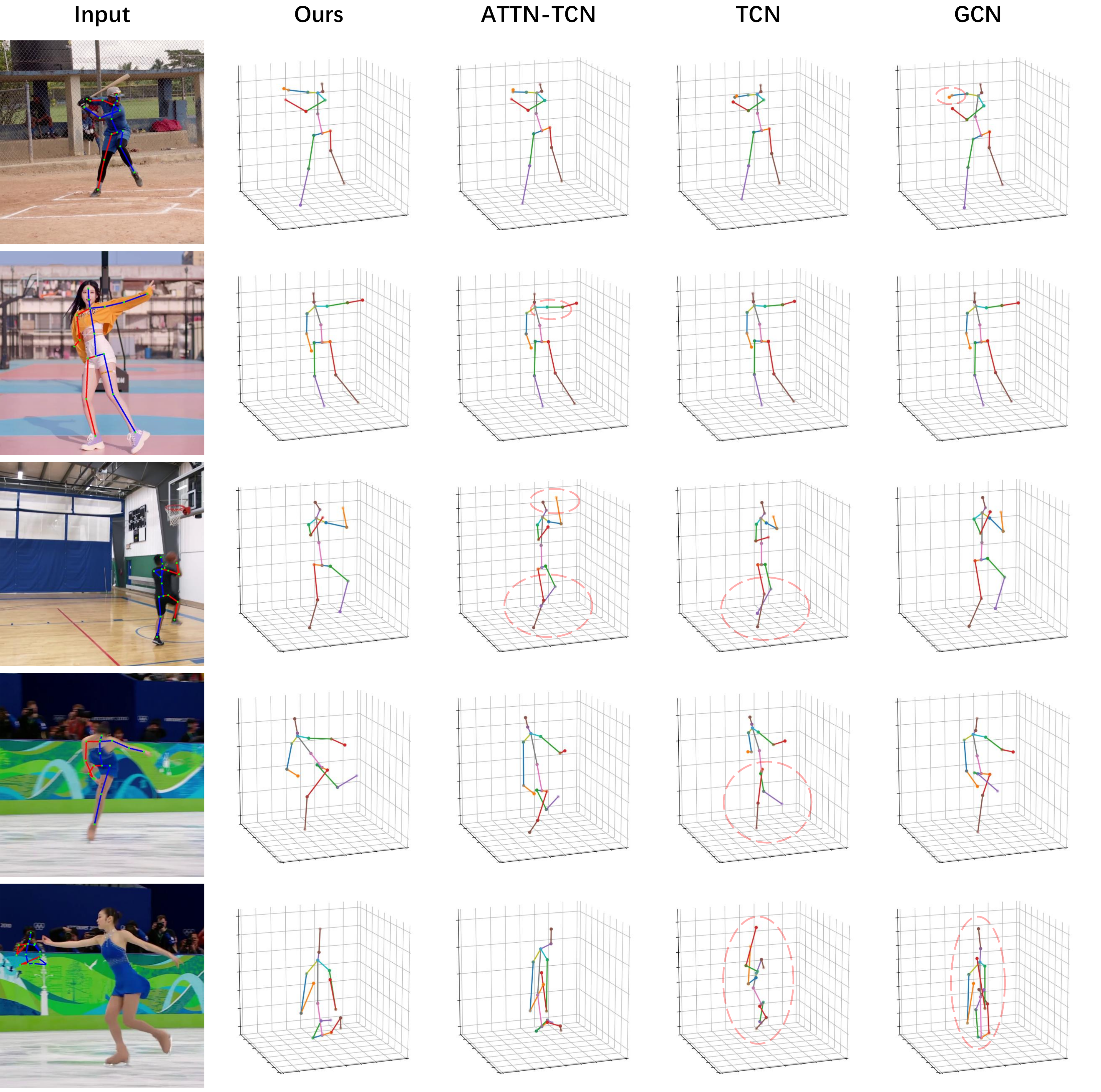}
   \caption
   {
      Qualitative comparisons on challenging in-the-wild videos with previous state-of-the-art methods, ATTN-TCN~\cite{liu2020attention}, TCN~\cite{pavllo20193d}, and GCN~\cite{cai2019exploiting}.  
      The last row shows the failure case, where the 2D detector has failed badly. 
   }
   \label{fig:wild_compare}
\end{figure}

\subsection{Qualitative Results}
\textbf{Attention visualization.}
Our method is easily interpretable through visualizing the attention score across frames to explain what the target frame relies on. 
Visualization results of the multi-head attention maps of the first attention layers from VTE and STE (243-frame model) are shown in Fig.~\ref{fig:attention}. 
The left map shows strong attention close to the input frames~\cite{Wu2020LiteTransformer,jiang2020convbert}, while the right map mainly pays strong attention to the center frame across all the sequences. 
This is expected since the proposed full-to-single strategy enables the VTE and STE modules to learn different representations: 
(i) VTE selectively identifies important sequences that are close to the input frames and enforces temporal consistency across frames. 
(ii) STE learns a specific representation from the input sequences using both past and future data, improving the representation ability of features to reach an optimal inference for the target frame. 
Note that a few attention head maps are sparse due to the different temporal patterns or semantics. 

\textbf{3D reconstruction visualization.}
We further evaluate our method on challenging in-the-wild videos from YouTube. 
Fig.~\ref{fig:wild_compare} shows the qualitative comparisons with the previous state-of-the-art methods~\cite{liu2020attention,pavllo20193d,cai2019exploiting}. 
We use the same 2D detector (cascaded pyramid network~\cite{chen2018cascaded}) to obtain 2D poses and then feed them to the models for a fair comparison. 
Despite the challenging samples with
complex actions and fast movements, the proposed method can produce realistic and structurally plausible 3D predictions outperforming previous works. 
This demonstrates our method is robust to partial occlusions and tolerant to depth ambiguity. 
The last row shows the failure case caused by a big 2D detection error.

\section{Conclusion} 
In this work, we investigate the suitableness of applying a Transformer-based network to the task of video-based 3D human pose estimation. 
From the proposed Strided Transformer with Strided Transformer Encoder (STE) and full-to-single supervision scheme, we show how the representative single-pose representation can be learned from redundant sequences. 
The key is to reasonably use strided convolutions in the Transformer architecture to aggregate long-range information into a single-vector pose in a hierarchical global and local fashion. 
Meanwhile, the computation cost can be reduced significantly. 
Moreover, our full-to-single supervision scheme enhances temporal smoothness and further refines the representation for the target frame. 
Comprehensive experiments on two benchmark datasets demonstrate that our method achieves superior performance compared with state-of-the-art methods. 

Although our method can reduce the computation cost of Transformers, the computational complexity and runtime cost of our method are still larger than temporal convolutional networks~\cite{pavllo20193d,chen2021anatomy}, indicated in Table~\ref{table:compare}. 
It is well acknowledged that the strong performance of Transformers comes at high computational costs. 
Note that the scope of this paper only targets improving FFN in the Transformer model. 
Future works may include designing a more efficient self-attention mechanism and extending our Strided Transformer to solve multi-view 3D human pose estimation. 
In addition, we hope that our approach would bring inspiration to the field of skeleton-based representation learning, \emph{e.g.}, action recognition, motion prediction, pose tracking, and so on. 

\bibliographystyle{IEEEtran}
\bibliography{ref.bib}

\begin{thebibliography}{10}
\providecommand{\url}[1]{#1}
\csname url@samestyle\endcsname
\providecommand{\newblock}{\relax}
\providecommand{\bibinfo}[2]{#2}
\providecommand{\BIBentrySTDinterwordspacing}{\spaceskip=0pt\relax}
\providecommand{\BIBentryALTinterwordstretchfactor}{4}
\providecommand{\BIBentryALTinterwordspacing}{\spaceskip=\fontdimen2\font plus
\BIBentryALTinterwordstretchfactor\fontdimen3\font minus
  \fontdimen4\font\relax}
\providecommand{\BIBforeignlanguage}[2]{{%
\expandafter\ifx\csname l@#1\endcsname\relax
\typeout{** WARNING: IEEEtran.bst: No hyphenation pattern has been}%
\typeout{** loaded for the language `#1'. Using the pattern for}%
\typeout{** the default language instead.}%
\else
\language=\csname l@#1\endcsname
\fi
#2}}
\providecommand{\BIBdecl}{\relax}
\BIBdecl

\bibitem{radwan2013monocular}
I.~Radwan, A.~Dhall, and R.~Goecke, ``Monocular image 3d human pose estimation
  under self-occlusion,'' in \emph{Proceedings of the IEEE International
  Conference on Computer Vision (ICCV)}, 2013, pp. 1888--1895.

\bibitem{li20143d}
S.~Li and A.~B. Chan, ``3d human pose estimation from monocular images with
  deep convolutional neural network,'' in \emph{Asian Conference on Computer
  Vision (ACCV)}, 2014, pp. 332--347.

\bibitem{zhao20183}
T.~Zhao, S.~Li, K.~N. Ngan, and F.~Wu, ``3-d reconstruction of human body shape
  from a single commodity depth camera,'' \emph{IEEE Transactions on
  Multimedia}, vol.~21, no.~1, pp. 114--123, 2018.

\bibitem{hu20213dbodynet}
P.~Hu, E.~S.-l. Ho, and A.~Munteanu, ``3dbodynet: Fast reconstruction of 3d
  animatable human body shape from a single commodity depth camera,''
  \emph{IEEE Transactions on Multimedia}, 2021.

\bibitem{kadkhodamohammadi2021generalizable}
A.~Kadkhodamohammadi and N.~Padoy, ``A generalizable approach for multi-view 3d
  human pose regression,'' \emph{Machine Vision and Applications}, vol.~32,
  no.~1, pp. 1--14, 2021.

\bibitem{pullen2002motion}
K.~Pullen and C.~Bregler, ``Motion capture assisted animation: Texturing and
  synthesis,'' in \emph{Proceedings of the 29th annual conference on Computer
  graphics and interactive techniques}, 2002, pp. 501--508.

\bibitem{wang2018depth}
P.~Wang, W.~Li, Z.~Gao, C.~Tang, and P.~O. Ogunbona, ``Depth pooling based
  large-scale 3-d action recognition with convolutional neural networks,''
  \emph{IEEE Transactions on Multimedia}, vol.~20, no.~5, pp. 1051--1061, 2018.

\bibitem{liu2017robust}
M.~Liu, H.~Liu, and C.~Chen, ``Robust 3d action recognition through sampling
  local appearances and global distributions,'' \emph{IEEE Transactions on
  Multimedia}, vol.~20, no.~8, pp. 1932--1947, 2017.

\bibitem{liu2018recognizing}
M.~Liu and J.~Yuan, ``Recognizing human actions as the evolution of pose
  estimation maps,'' in \emph{Proceedings of the IEEE Conference on Computer
  Vision and Pattern Recognition (CVPR)}, 2018, pp. 1159--1168.

\bibitem{wei2019learning}
P.~Wei, H.~Sun, and N.~Zheng, ``Learning composite latent structures for 3d
  human action representation and recognition,'' \emph{IEEE Transactions on
  Multimedia}, vol.~21, no.~9, pp. 2195--2208, 2019.

\bibitem{song2021constructing}
Y.-F. Song, Z.~Zhang, C.~Shan, and L.~Wang, ``Constructing stronger and faster
  baselines for skeleton-based action recognition,'' \emph{arXiv preprint
  arXiv:2106.15125}, 2021.

\bibitem{chen2021learning}
T.~Chen, D.~Zhou, J.~Wang, S.~Wang, Y.~Guan, X.~He, and E.~Ding, ``Learning
  multi-granular spatio-temporal graph network for skeleton-based action
  recognition,'' in \emph{Proceedings of the 29th ACM International Conference
  on Multimedia (ACMMM)}, 2021, pp. 4334--4342.

\bibitem{li2021memory}
C.~Li, C.~Xie, B.~Zhang, J.~Han, X.~Zhen, and J.~Chen, ``Memory attention
  networks for skeleton-based action recognition,'' \emph{IEEE Transactions on
  Neural Networks and Learning Systems}, 2021.

\bibitem{yang2021unik}
D.~Yang, Y.~Wang, A.~Dantcheva, L.~Garattoni, G.~Francesca, and F.~Bremond,
  ``Unik: A unified framework for real-world skeleton-based action
  recognition,'' \emph{arXiv preprint arXiv:2107.08580}, 2021.

\bibitem{zhang2017action}
B.~Zhang, Y.~Yang, C.~Chen, L.~Yang, J.~Han, and L.~Shao, ``Action recognition
  using 3d histograms of texture and a multi-class boosting classifier,''
  \emph{IEEE Transactions on Image processing}, vol.~26, no.~10, pp.
  4648--4660, 2017.

\bibitem{chen2017multi}
C.~Chen, M.~Liu, H.~Liu, B.~Zhang, J.~Han, and N.~Kehtarnavaz, ``Multi-temporal
  depth motion maps-based local binary patterns for 3-d human action
  recognition,'' \emph{IEEE Access}, vol.~5, pp. 22\,590--22\,604, 2017.

\bibitem{garcia2019human}
M.~Garcia-Salguero, J.~Gonzalez-Jimenez, and F.-A. Moreno, ``Human 3d pose
  estimation with a tilting camera for social mobile robot interaction,''
  \emph{Sensors}, vol.~19, no.~22, p. 4943, 2019.

\bibitem{gui2018teaching}
L.~Gui, K.~Zhang, Y.~Wang, X.~Liang, J.~M. Moura, and M.~Veloso, ``Teaching
  robots to predict human motion,'' in \emph{Proceedings of the IEEE
  International Conference on Intelligent Robots and Systems (IROS)}, 2018, pp.
  562--567.

\bibitem{martinez2017simple}
J.~Martinez, R.~Hossain, J.~Romero, and J.~J. Little, ``A simple yet effective
  baseline for 3d human pose estimation,'' in \emph{Proceedings of the IEEE
  International Conference on Computer Vision (ICCV)}, 2017, pp. 2640--2649.

\bibitem{pavllo20193d}
D.~Pavllo, C.~Feichtenhofer, D.~Grangier, and M.~Auli, ``3d human pose
  estimation in video with temporal convolutions and semi-supervised
  training,'' in \emph{Proceedings of the IEEE Conference on Computer Vision
  and Pattern Recognition (CVPR)}, 2019, pp. 7753--7762.

\bibitem{hua2021weakly}
G.~Hua, W.~Li, Q.~Zhang, R.~Ding, and H.~Liu, ``Weakly-supervised cross-view 3d
  human pose estimation,'' \emph{arXiv preprint arXiv:2105.10882}, 2021.

\bibitem{lee2018propagating}
K.~Lee, I.~Lee, and S.~Lee, ``Propagating lstm: 3d pose estimation based on
  joint interdependency,'' in \emph{Proceedings of the European Conference on
  Computer Vision (ECCV)}, 2018, pp. 119--135.

\bibitem{rayat2018exploiting}
M.~Rayat Imtiaz~Hossain and J.~J. Little, ``Exploiting temporal information for
  3d human pose estimation,'' in \emph{Proceedings of the European Conference
  on Computer Vision (ECCV)}, 2018, pp. 68--84.

\bibitem{cai2019exploiting}
Y.~Cai, L.~Ge, J.~Liu, J.~Cai, T.-J. Cham, J.~Yuan, and N.~M. Thalmann,
  ``Exploiting spatial-temporal relationships for 3d pose estimation via graph
  convolutional networks,'' in \emph{Proceedings of the IEEE International
  Conference on Computer Vision (ICCV)}, 2019, pp. 2272--2281.

\bibitem{Attention}
A.~Vaswani, N.~Shazeer, N.~Parmar, J.~Uszkoreit, L.~Jones, A.~N. Gomez, L.~u.
  Kaiser, and I.~Polosukhin, ``Attention is all you need,'' in \emph{Advances
  in Neural Information Processing Systems (NIPS)}, 2017, pp. 5998--6008.

\bibitem{tay2020efficient}
Y.~Tay, M.~Dehghani, D.~Bahri, and D.~Metzler, ``Efficient transformers: A
  survey,'' \emph{arXiv preprint arXiv:2009.06732}, 2020.

\bibitem{zihang2020funnel-transformer}
D.~Zihang, L.~Guokun, Y.~Yiming, and Q.~L. V., ``Funnel-transformer: Filtering
  out sequential redundancy for efficient language processing,'' in
  \emph{Advances in Neural Information Processing Systems (NIPS)}, 2020.

\bibitem{han2020survey}
K.~Han, Y.~Wang, H.~Chen, X.~Chen, J.~Guo, Z.~Liu, Y.~Tang, A.~Xiao, C.~Xu,
  Y.~Xu \emph{et~al.}, ``A survey on visual transformer,'' \emph{arXiv preprint
  arXiv:2012.12556}, 2020.

\bibitem{he2021transreid}
S.~He, H.~Luo, P.~Wang, F.~Wang, H.~Li, and W.~Jiang, ``Transreid:
  Transformer-based object re-identification,'' \emph{arXiv preprint
  arXiv:2102.04378}, 2021.

\bibitem{li2021trear}
X.~Li, Y.~Hou, P.~Wang, Z.~Gao, M.~Xu, and W.~Li, ``Trear: Transformer-based
  rgb-d egocentric action recognition,'' \emph{arXiv preprint
  arXiv:2101.03904}.

\bibitem{han2020exploiting}
L.~Han, P.~Wang, Z.~Yin, F.~Wang, and H.~Li, ``Exploiting better feature
  aggregation for video object detection,'' in \emph{Proceedings of the 28th
  ACM International Conference on Multimedia (ACMMM)}, 2020, pp. 1469--1477.

\bibitem{li2021transformer}
X.~Li, Y.~Hou, P.~Wang, Z.~Gao, M.~Xu, and W.~Li, ``Transformer guided geometry
  model for flow-based unsupervised visual odometry,'' \emph{Neural Computing
  and Applications}, pp. 1--12, 2021.

\bibitem{geva2020transformer}
M.~Geva, R.~Schuster, J.~Berant, and O.~Levy, ``Transformer feed-forward layers
  are key-value memories,'' \emph{arXiv preprint arXiv:2012.14913}, 2020.

\bibitem{liu2020attention}
R.~Liu, J.~Shen, H.~Wang, C.~Chen, S.-c. Cheung, and V.~Asari, ``Attention
  mechanism exploits temporal contexts: Real-time 3d human pose
  reconstruction,'' in \emph{Proceedings of the IEEE Conference on Computer
  Vision and Pattern Recognition (CVPR)}, 2020, pp. 5064--5073.

\bibitem{ionescu2013human3}
C.~Ionescu, D.~Papava, V.~Olaru, and C.~Sminchisescu, ``Human3.6m: Large scale
  datasets and predictive methods for 3d human sensing in natural
  environments,'' \emph{IEEE Transactions on Pattern Analysis and Machine
  Intelligence}, vol.~36, no.~7, pp. 1325--1339, 2013.

\bibitem{sigal2010humaneva}
L.~Sigal, A.~O. Balan, and M.~J. Black, ``Humaneva: Synchronized video and
  motion capture dataset and baseline algorithm for evaluation of articulated
  human motion,'' \emph{International Journal of Computer Vision}, vol.~87,
  no.~12, pp. 4--27, 2010.

\bibitem{pavlakos2017coarse}
G.~Pavlakos, X.~Zhou, K.~G. Derpanis, and K.~Daniilidis, ``Coarse-to-fine
  volumetric prediction for single-image 3d human pose,'' in \emph{Proceedings
  of the IEEE Conference on Computer Vision and Pattern Recognition (CVPR)},
  2017, pp. 7025--7034.

\bibitem{sun2018integral}
X.~Sun, B.~Xiao, F.~Wei, S.~Liang, and Y.~Wei, ``Integral human pose
  regression,'' in \emph{Proceedings of the European Conference on Computer
  Vision (ECCV)}, 2018, pp. 529--545.

\bibitem{zhao2019semantic}
L.~Zhao, X.~Peng, Y.~Tian, M.~Kapadia, and D.~N. Metaxas, ``Semantic graph
  convolutional networks for 3d human pose regression,'' in \emph{Proceedings
  of the IEEE Conference on Computer Vision and Pattern Recognition (CVPR)},
  2019, pp. 3425--3435.

\bibitem{liu2019feature}
J.~Liu, H.~Ding, A.~Shahroudy, L.-Y. Duan, X.~Jiang, G.~Wang, and A.~C. Kot,
  ``Feature boosting network for 3d pose estimation,'' \emph{IEEE Transactions
  on Pattern Analysis and Machine Intelligence}, vol.~42, no.~2, pp. 494--501,
  2019.

\bibitem{fang2018learning}
H.-S. Fang, Y.~Xu, W.~Wang, X.~Liu, and S.-C. Zhu, ``Learning pose grammar to
  encode human body configuration for 3d pose estimation,'' in
  \emph{Thirty-Second AAAI Conference on Artificial Intelligence}, 2018.

\bibitem{xu2021graph}
T.~Xu and W.~Takano, ``Graph stacked hourglass networks for 3d human pose
  estimation,'' in \emph{Proceedings of the IEEE Conference on Computer Vision
  and Pattern Recognition (CVPR)}, 2021, pp. 16\,105--16\,114.

\bibitem{gong2021poseaug}
K.~Gong, J.~Zhang, and J.~Feng, ``Poseaug: A differentiable pose augmentation
  framework for 3d human pose estimation,'' in \emph{Proceedings of the IEEE
  Conference on Computer Vision and Pattern Recognition (CVPR)}, 2021, pp.
  8575--8584.

\bibitem{wang2020motion}
J.~Wang, S.~Yan, Y.~Xiong, and D.~Lin, ``Motion guided 3d pose estimation from
  videos,'' \emph{arXiv preprint arXiv:2004.13985}, 2020.

\bibitem{chen2021anatomy}
T.~Chen, C.~Fang, X.~Shen, Y.~Zhu, Z.~Chen, and J.~Luo, ``Anatomy-aware 3d
  human pose estimation with bone-based pose decomposition,'' \emph{IEEE
  Transactions on Circuits and Systems for Video Technology}, 2021.

\bibitem{carion2020end}
N.~Carion, F.~Massa, G.~Synnaeve, N.~Usunier, A.~Kirillov, and S.~Zagoruyko,
  ``End-to-end object detection with transformers,'' in \emph{European
  Conference on Computer Vision (ECCV)}, 2020, pp. 213--229.

\bibitem{zhu2020deformable}
X.~Zhu, W.~Su, L.~Lu, B.~Li, X.~Wang, and J.~Dai, ``Deformable detr: Deformable
  transformers for end-to-end object detection,'' \emph{arXiv preprint
  arXiv:2010.04159}, 2020.

\bibitem{dosovitskiy2020image}
A.~Dosovitskiy, L.~Beyer, A.~Kolesnikov, D.~Weissenborn, X.~Zhai,
  T.~Unterthiner, M.~Dehghani, M.~Minderer, G.~Heigold, S.~Gelly \emph{et~al.},
  ``An image is worth 16x16 words: Transformers for image recognition at
  scale,'' \emph{arXiv preprint arXiv:2010.11929}, 2020.

\bibitem{yuan2021tokens}
L.~Yuan, Y.~Chen, T.~Wang, W.~Yu, Y.~Shi, F.~E. Tay, J.~Feng, and S.~Yan,
  ``Tokens-to-token vit: Training vision transformers from scratch on
  imagenet,'' \emph{arXiv preprint arXiv:2101.11986}, 2021.

\bibitem{lin2020end}
K.~Lin, L.~Wang, and Z.~Liu, ``End-to-end human pose and mesh reconstruction
  with transformers,'' \emph{arXiv preprint arXiv:2012.09760}, 2020.

\bibitem{lin2019trajectory}
J.~Lin and G.~H. Lee, ``Trajectory space factorization for deep video-based 3d
  human pose estimation,'' \emph{arXiv preprint arXiv:1908.08289}, 2019.

\bibitem{xu2020deep}
J.~Xu, Z.~Yu, B.~Ni, J.~Yang, X.~Yang, and W.~Zhang, ``Deep kinematics analysis
  for monocular 3d human pose estimation,'' in \emph{Proceedings of the IEEE
  Conference on Computer Vision and Pattern Recognition (CVPR)}, 2020, pp.
  899--908.

\bibitem{zeng2020srnet}
A.~Zeng, X.~Sun, F.~Huang, M.~Liu, Q.~Xu, and S.~Lin, ``Srnet: Improving
  generalization in 3d human pose estimation with a split-and-recombine
  approach,'' in \emph{European Conference on Computer Vision (ECCV)}, 2020,
  pp. 507--523.

\bibitem{pavlakos2018ordinal}
G.~Pavlakos, X.~Zhou, and K.~Daniilidis, ``Ordinal depth supervision for 3d
  human pose estimation,'' in \emph{Proceedings of the IEEE Conference on
  Computer Vision and Pattern Recognition (CVPR)}, 2018, pp. 7307--7316.

\bibitem{liu2020comprehensive}
K.~Liu, R.~Ding, Z.~Zou, L.~Wang, and W.~Tang, ``A comprehensive study of
  weight sharing in graph networks for 3d human pose estimation,'' in
  \emph{Proceedings of the European Conference on Computer Vision (ECCV)},
  2020, pp. 318--334.

\bibitem{chen2019weakly}
X.~Chen, K.-Y. Lin, W.~Liu, C.~Qian, and L.~Lin, ``Weakly-supervised discovery
  of geometry-aware representation for 3d human pose estimation,'' in
  \emph{Proceedings of the IEEE Conference on Computer Vision and Pattern
  Recognition (CVPR)}, 2019, pp. 10\,895--10\,904.

\bibitem{tome2018rethinking}
D.~Tome, M.~Toso, L.~Agapito, and C.~Russell, ``Rethinking pose in 3d:
  Multi-stage refinement and recovery for markerless motion capture,'' in
  \emph{2018 International Conference on 3D Vision (3DV)}, 2018, pp. 474--483.

\bibitem{kocabas2019self}
M.~Kocabas, S.~Karagoz, and E.~Akbas, ``Self-supervised learning of 3d human
  pose using multi-view geometry,'' in \emph{Proceedings of the IEEE Conference
  on Computer Vision and Pattern Recognition (CVPR)}, 2019, pp. 1077--1086.

\bibitem{cheng2019occlusion}
Y.~Cheng, B.~Yang, B.~Wang, W.~Yan, and R.~T. Tan, ``Occlusion-aware networks
  for 3d human pose estimation in video,'' in \emph{Proceedings of the IEEE
  International Conference on Computer Vision (ICCV)}, 2019, pp. 723--732.

\bibitem{chen2018cascaded}
Y.~Chen, Z.~Wang, Y.~Peng, Z.~Zhang, G.~Yu, and J.~Sun, ``Cascaded pyramid
  network for multi-person pose estimation,'' in \emph{Proceedings of the IEEE
  Conference on Computer Vision and Pattern Recognition (CVPR)}, 2018, pp.
  7103--7112.

\bibitem{he2017mask}
K.~He, G.~Gkioxari, P.~Doll{\'a}r, and R.~Girshick, ``Mask r-cnn,'' in
  \emph{Proceedings of the IEEE international conference on computer vision
  (ICCV)}, 2017, pp. 2961--2969.

\bibitem{dabral2018learning}
R.~Dabral, A.~Mundhada, U.~Kusupati, S.~Afaque, A.~Sharma, and A.~Jain,
  ``Learning 3d human pose from structure and motion,'' in \emph{Proceedings of
  the European Conference on Computer Vision (ECCV)}, 2018, pp. 668--683.

\bibitem{newell2016stacked}
A.~Newell, K.~Yang, and J.~Deng, ``Stacked hourglass networks for human pose
  estimation,'' in \emph{Proceedings of the European Conference on Computer
  Vision (ECCV)}, 2016, pp. 483--499.

\bibitem{Wu2020LiteTransformer}
Z.~Wu, Z.~Liu, J.~Lin, Y.~Lin, and S.~Han, ``Lite transformer with long-short
  range attention,'' in \emph{International Conference on Learning
  Representations (ICLR)}, 2020.

\bibitem{jiang2020convbert}
Z.~Jiang, W.~Yu, D.~Zhou, Y.~Chen, J.~Feng, and S.~Yan, ``Convbert: Improving
  bert with span-based dynamic convolution,'' in \emph{Advances in Neural
  Information Processing Systems (NIPS)}, 2020.

\end{thebibliography}

\newpage
\section{Appendix} 

\begin{figure*}[!hb]
	\centering
	\includegraphics[width=0.62 \linewidth]{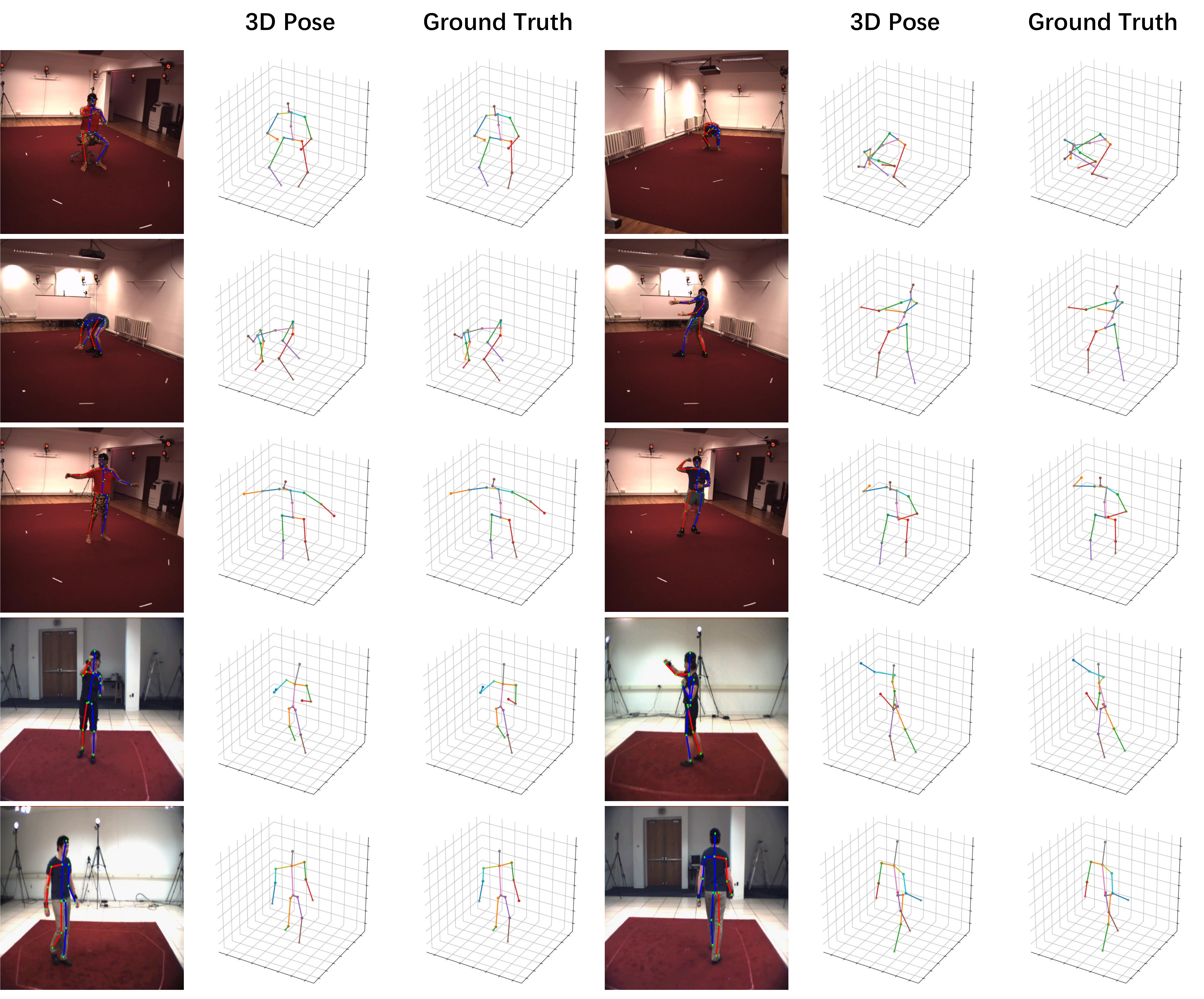}
	\caption
	{
    Visual results of our proposed method on Human3.6M dataset (first 3 rows) and HumanEva-I dataset (last 2 rows). 
	}
	\label{fig:dataset}
\end{figure*}

\begin{figure*}[!hb]
	\centering
	\includegraphics[width=0.86 \linewidth]{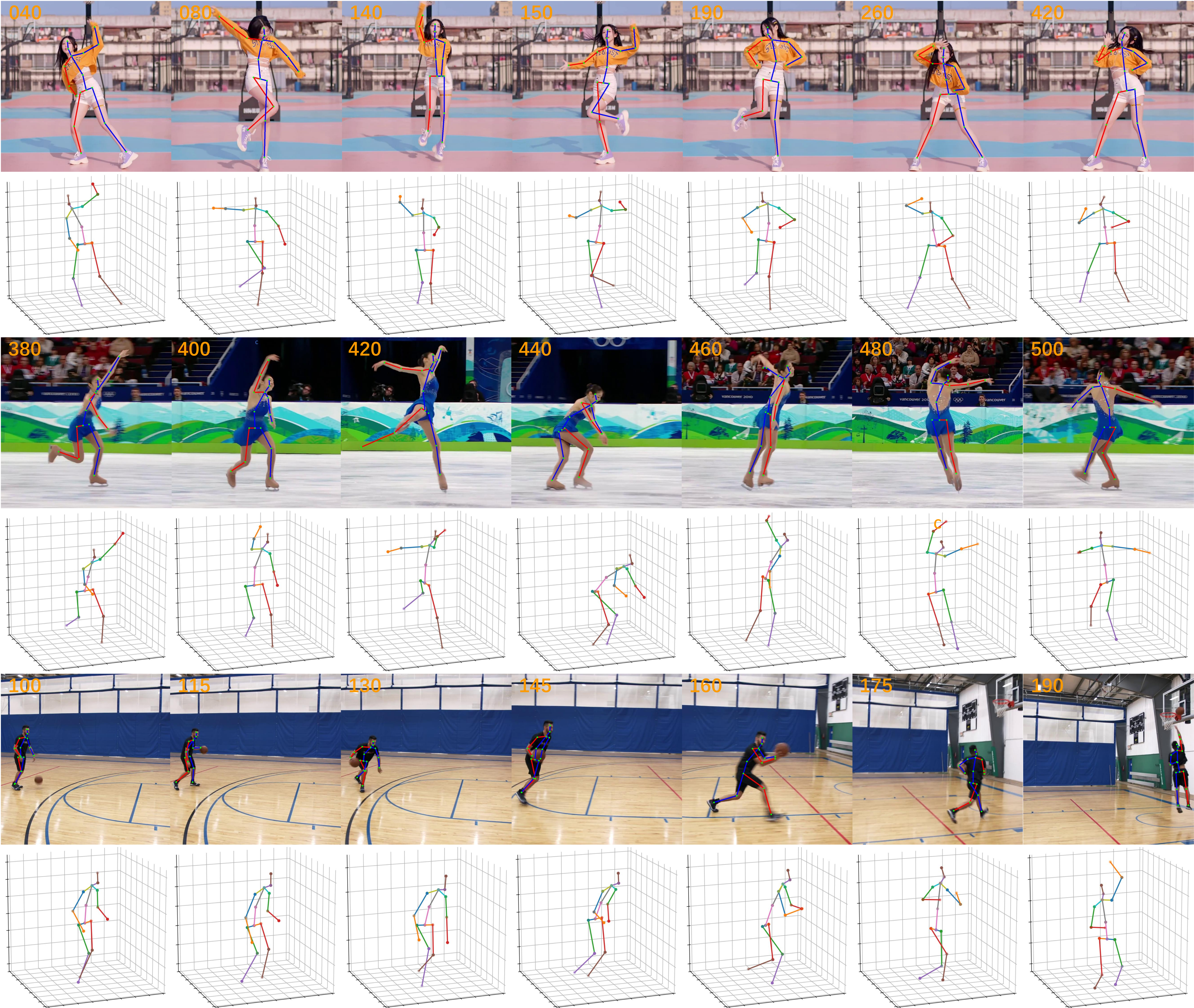}
	\caption
	{
    Qualitative results on challenging wild videos. 
    The number is the frame index of input videos.
	}
	\label{fig:wild}
\end{figure*}

\end{document}